\documentclass[a4paper,fleqn]{cas-dc}
\usepackage{amssymb}
\usepackage[numbers]{natbib}
\usepackage{subfig}
\usepackage{siunitx}
\usepackage{graphicx}
\usepackage{float}
\usepackage{amsmath}
\usepackage{mathtools}
\usepackage{multicol,lipsum}
\usepackage[justification=centering]{caption}

\usepackage{cuted}
\usepackage{flushend}

\newcommand*\rot{\rotatebox{90}}
\DeclarePairedDelimiter{\ceil}{\lceil}{\rceil}
\newcommand\eatpunct[1]{}
\def\tsc#1{\csdef{#1}{\textsc{\lowercase{#1}}\xspace}}
\tsc{WGM}
\tsc{QE}
\tsc{EP}
\tsc{PMS}
\tsc{BEC}
\tsc{DE}

\begin{document}
\let\WriteBookmarks\relax
\def\floatpagepagefraction{1}
\def\textpagefraction{.001}
\shorttitle{Pre-event and post-event information fusion}
\shortauthors{Lenjani et~al.}

\title [mode = title]{Towards fully automated post-event data collection and analysis: pre-event and post-event information fusion}                      

\author[1]{\textcolor{Black}{Ali Lenjani}}
\cormark[1]
\ead{alenjani@purdue.edu}
\author[1,2]{\textcolor{Black}{Shirley J. Dyke}}
\author[1]{\textcolor{Black}{Ilias Bilionis}}
\author[3]{\textcolor{Black}{Chul Min Yeum}}
\author[1]{\textcolor{Black}{Kenzo Kamiya}}
\author[1]{\textcolor{Black}{Jongseong Choi}}
\author[1]{\textcolor{Black}{Xiaoyu Liu}}
\author[4]{\textcolor{Black}{Arindam G. Chowdhury}}

\address[1]{School of Mechanical Engineering, Purdue University, West Lafayette, IN, USA}
\address[2]{Lyles School of Civil Engineering, Purdue University, West Lafayette, IN, USA}
\address[3]{Department of Civil and Environmental Engineering, University of Waterloo, ON, N2L 3G1, Canada}
\address[4]{Department of Civil and Environmental Engineering, Florida International University, Miami, FL, USA}

\begin{abstract}
In post-event reconnaissance missions, engineers and researchers collect perishable information about damaged buildings in the affected geographical region to learn from the consequences of the event. 
A typical post-event reconnaissance mission is conducted by first doing a preliminary survey, followed by a detailed survey. 
The objective of the preliminary survey is to develop an understanding of the overall situation in the field, and use that information to plan the detailed survey.
The preliminary survey is typically conducted by driving slowly along a pre-determined route, observing the damage, and noting where further detailed data should be collected. 
This involves several manual, time-consuming steps that can be accelerated by exploiting recent advances in computer vision and artificial intelligence.
The objective of this work is to develop and validate an automated technique to support post-event reconnaissance teams in the rapid collection of reliable and sufficiently comprehensive data, for planning the detailed survey. The focus here is on residential buildings. 
The technique incorporates several methods designed to automate the process of categorizing buildings based on their key physical attributes, and rapidly assessing their post-event structural condition.
It is divided into pre-event and post-event streams, each intending to first extract all possible information about the target buildings using both pre-event and post-event images.
Algorithms based on convolutional neural network (CNNs) are implemented for scene (image) classification.
A probabilistic approach is developed to fuse the results obtained from analyzing several images to yield a robust decision regarding the attributes and condition of a target building. 
We validate the technique using post-event images captured during reconnaissance missions that took place after hurricanes Harvey and Irma. The validation data were collected by a structural wind and coastal engineering reconnaissance team, the National Science Foundation (NSF) funded Structural Extreme Events Reconnaissance (StEER) Network.
\end{abstract}




\maketitle

\section{Introduction}
\label{sec:intro}
Rapid reconnaissance teams have been deployed after significant natural hazard events for decades with the objective of collecting perishable information to be used by scientists and engineers to learn from the event consequences. 
Such data have been instrumental in revealing gaps in knowledge, improving design procedures and building codes, and generally reducing the vulnerability of the built environment. 
There has been an enormous investment directed toward the collection of these data, based on the expectation that these data will be even more critical in the future. 
For example, in the United States, the Natural Hazards Engineering Research Infrastructure (NHERI), a distributed network funded by the National Science Foundation \cite{nsfgrant2015nheri,allnsfgranttil2019nheri,nsfgrant2019nheri}, includes the Post-Disaster, Rapid Response Research (RAPID) Facility to support data collection and use \cite{nsfgrantrapid,rapid}. 
NHERI has developed a Science Plan to guide scientific efforts, which stresses the need to better collect and share data and information to enable research and deliver solutions \cite{nheriscienceplan}. 
The NHERI Science Plan also emphasizes the need to collect and analyze sensor and image information for use in disaster preparedness, mitigation, response, and recovery.
The most recent addition to the NHERI network is the CONVERGE center, headquartered at the University of Colorado at Boulder, which aims to coordinate hazards and disaster researchers to better link them to NHERI partners \cite{nsfgrant2019converge,converge}. 
CONVERGE anticipates leveraging and advancing the platforms, networks, mobile applications, cyberinfrastructure, and research opportunities for these reconnaissance teams to leverage. 
One of the key partners leading the structural engineering data collection efforts is the Structural Extreme Events Reconnaissance (StEER) Network \cite{nsfgrantsteer,steer}. 
In addition, the Earthquake Engineering Research Institute (EERI) also initiated the Virtual Earthquake Reconnaissance Team (VERT) that aims to engage young engineers and graduate students in post-disaster reconnaissance \cite{vert}.

The data collection platforms that support these efforts, including drones and satellites, have advanced rapidly in recent years. 
However, many of the steps involved in the organization and analysis of the complex and unstructured data collected during post-event reconnaissance missions are still predominantly manual and quite time-consuming. 
Furthermore, the research needed to accelerate, and even automate, the analysis of these data has not kept pace with the enormous investment directed toward the collection of these data. 
Automating some of the procedures associated with building damage surveys will enable reconnaissance teams to more rapidly gather and analyze these large volumes of perishable information. 
Recent demonstrations of automation include scene recognition and object detection with large volumes of images collected after an event by exploiting new developments in convolutional neural networks (CNNs) \cite{choi2018computer,gao2018deep,yeum2018postevent,yeum2018visual}. 
These techniques, which fall into the broad category of artificial intelligence, are gaining traction. 
However, there are still significant challenges associated with real world application of these methods, mainly revolving around both the need to acquire sufficient quantities of ground truth data and the potential to inadvertently introduce bias into the training process \cite{everingham2010pascal}. 

Here we develop an end-to-end technique for automating several steps in the analysis and decisions associated with post-event damage survey data. 
Post-event surveys can be broken down into a preliminary survey, sometimes called a ``windshield survey,'' followed by a detailed survey \cite{comerio1998disaster}. 
The preliminary survey is conducted to collect initial data to gain a perspective about the overall situation in the field. This initial data are then used to make decisions regarding what further data must be collected during the detailed survey. 
To conduct the preliminary survey, field engineers usually drive slowly along the streets in the affected region to observe the extent of the damage. 
This typically takes place within a few days of the event. 
These coarse data might be augmented by occasionally getting out of the vehicle to take photos or perhaps to get a closer look at debris or specific buildings. 
The preliminary survey is conducted to provide evidence that is used to plan an efficient detailed survey.
During the detailed survey, several small teams of engineers and architects, data collectors, are dispatched to the region to visit specific buildings and collect much more detailed information about their condition \cite{designsafe,geer,steer}. 
Typically, the detailed survey involves collecting these data by walking around each building, or even entering the building if permitted to do so.
Many of these teams intend to capture data that may motivate new lines of scientific inquiry related to the performance of our infrastructure. 

Within our procedure we also leverage relatively new vision sensors, such as spherical cameras that can be mounted on street view cars, that have the mobility to rapidly collect a large volume of entire-view, high-resolution images in a short period of time \cite{anguelov2010google}. 
To support many other needs in the commercial sector, regularly-updated images of buildings' facades are captured and stored through street view services. 
These images may be critical for damage surveys, as after an event a building may be so severely damaged that its original attributes may not be decipherable. 
An automated technique has been developed to extract high-quality pre-event images from several viewpoints using only a single geo-tagged image or its GPS data\cite{lenjani2019automated}.
Additionally, after the event, images may be similarly collected with spherical cameras to quickly record the external appearance of buildings and support visual assessment\cite{lenjani2019automated,yeum2018automated}. 
The integration of these readily available data, efficient and automated analytics capabilities, and processing power, can greatly improve the efficiency of the reconnaissance missions.

The objective of this research is to develop and validate an automated technique to process post-event reconnaissance image data and output the relevant attributes and overall damage condition of each building.
Using only the visual content in the images, the technique is intended to directly support engineers and architects mainly during the preliminary survey phase of a reconnaissance mission. 
Automation is applied to extract the relevant information typically collected during such missions, making it readily available to the human engineer and architect that must act upon that information. 
We first develop an appropriate classification schema for this application and establish the ability to categorize buildings based on their key physical attributes using pre-event data. 
CNNs are utilized for scene (image) classification to categorize the target building, shown in a set of images, based on their structural attributes and post-event condition.
Next, post-event data is similarly used to rapidly determine their post-event condition.
In each case, by appropriately fusing the information extracted from multiple images, we make robust determinations regarding the categorization of each building. 

The information fusion process developed and integrated into the technique considers the quality and completeness of the data collected. 
We validate the technique using post-event images of residential buildings captured during hurricane Harvey and Irma reconnaissance missions collected by the NSF-funded StEER Network \cite{steer_data,steer}. 
We evaluate the performance of the technique by comparing our results to the documentation collected during the mission, as recorded through the Fulcrum app \cite{fulcrum}, and we discuss the need for greater volumes of data to be collected in future missions. 

The remainder of this paper is organized as follows: 
Sec. \ref{sec:method} provides the problem formulation. 
Sec. \ref{sec:exp_val} provides a demonstration and validation of its effectiveness. 
The conclusions are discussed in Sec. \ref{sec:con}.

\section{Technical approach}
\label{sec:method}

\begin{figure*}
     \centering
     \includegraphics[width=1\textwidth]{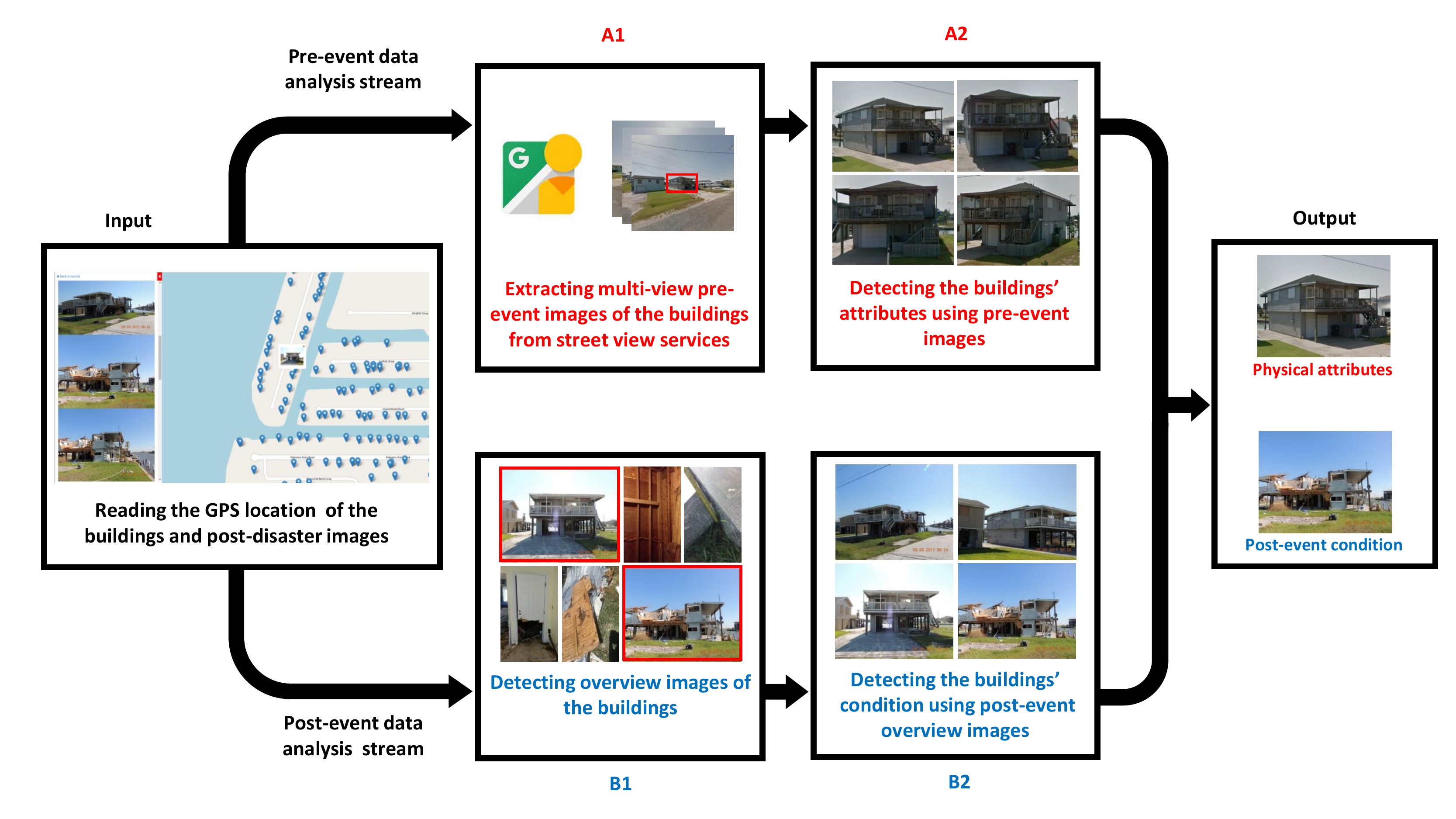}
     \caption{Diagram showing the steps in the automated procedure.}
     \label{fig:technique}
\end{figure*}

A general diagram of the technique developed is shown in, Fig.\ref{fig:technique}. 
The input is a collection of geo-tagged, post-event images of the residential buildings in a region. 
The output is the information needed for an assessment of each residential building, including automatically generated physical and structural attributes plus post-event condition information. 
Certain necessary physical and structural attributes are best obtained from the pre-event condition, so multiple pre-event images are automatically extracted from existing street view databases. 
Post-event building condition information is obtained directly from post-event images. 

The technique is implemented through two branches of data analysis, conducted independently.
We call these two branches the \emph{post-event data analysis stream} and the \emph{pre-event data analysis stream}.
The post-event stream detects assesses the overall damage condition of the building after the event based on the images collected during the preliminary survey. 
The pre-event stream extracts building physical attributes to be used for the preliminary screening, as well as several pre-event views of the building from various perspectives. 
These two sets of complementary information are organized in a way that assists the decision-making process of human inspectors regarding where to focus resources during a detailed survey. 
For clarity, we design a classification schema specific to post-event preliminary surveys.
The schema can be easily extended to support other applications. 
In the subsequent paragraphs, we discuss the process use to develop each data analysis stream.
The detailed definitions for the classification schema are provided in Sec. \ref{sec:cls_schm}.

The post-event data analysis stream requires the design and training of two image classifiers which are implemented sequentially.
The first classifier is intended to filter out images that contain useful information about the condition of the building, step \emph{B1}. 
The best images for detecting the overall condition of the building for hurricane assessment are images that provide a view of the entire building. 
However, the data collected for a given target building may include close images of components or details, or even irrelevant images (e.g., cars, trees, windows, doors, etc). 
Including these in the dataset to be automatically analyzed may bias the results, or increase the processing time.
The filtered data are passed to the next classifier, which is trained to detect the overall condition of the structure, step \emph{B2}, see Sec. \ref{sec:cls_schm_post}. 

The pre-event data analysis stream automatically detects certain physical attributes of each building that are useful in a preliminary post-event survey using image classification.
Since post-event images of buildings that have experienced severe damage cannot reliably be used to determine the original physical attributes, it is more appropriate to use pre-event images for this purpose. 
To this end, we developed a fully automated technique to extract pre-event images from street view imagery services, step \emph{A1}.
These pre-event images along with the ground truth labels, provided by the field engineers \cite{steer_data}, are used to design and train a set of image classifiers, that can detect certain physical attributes, explained in Sec. \ref{sec:cls_schm_pre}, step \emph{A2}. 

In some cases, reliable determination of a physical attribute or even the condition of the building requires that classification results from several images containing multiple views of the building be used. 
For instance, if several post-event images are collected from a building, and only one of those images provides a view of the damaged region, the classifier will only detect damage in that one specific image; The specific image containing the damage cannot be known in advance. 
Therefore, the relevant images available must be used collectively to make a determination. 
We have developed an approach to fuse the information from several images to make such decisions. 
The problem formulation is provided in Sec. \ref{sec:info_fuse} and the demonstration is included in Sec. \ref{sec:exp_val}.

\subsection{Design of the classification schema}
\label{sec:cls_schm}

\begin{figure*}
	\centering
	\subfloat[ \text{Pre-event}
	\label{fig:hir_pre}]{\includegraphics[scale= .5]{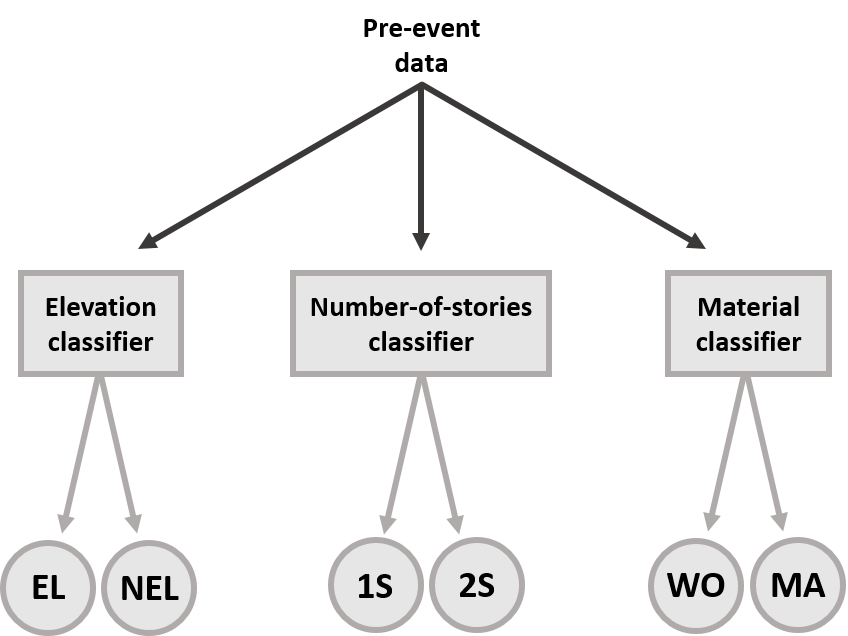}}
    \hspace{40pt}
	\subfloat[\text{Post-event} \label{fig:hir_post}]{\includegraphics[scale= 0.5]{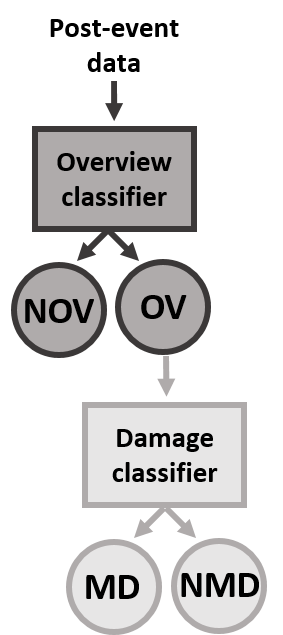}}
	\caption{Hierarchy of classifiers used in \text{pre-event} and \text{post-event} data analysis streams.}
	\label{fig:hir}
\end{figure*}

The classification schema designed to support preliminary hurricane surveys is shown in, Fig. \ref{fig:hir}, (the abbreviations are defined later).
Classifiers are much more effective when clear boundaries exist to distinguish the visual features of the images in different classes. 
This is especially true to achieve robust classification in the real world when using such unstructured and complex data, as is often the case in reconnaissance datasets. 
Thus, a clear definition for each class is needed to establish consistent ground-truth data that are suitable for training. 
The definitions for those comprising the post-event and pre-event streams are discussed in the following sections. 

\subsubsection{Classifiers used in the post-event stream}
\label{sec:cls_schm_post}
\begin{figure*}
     \centering
     \includegraphics[width=1\textwidth]{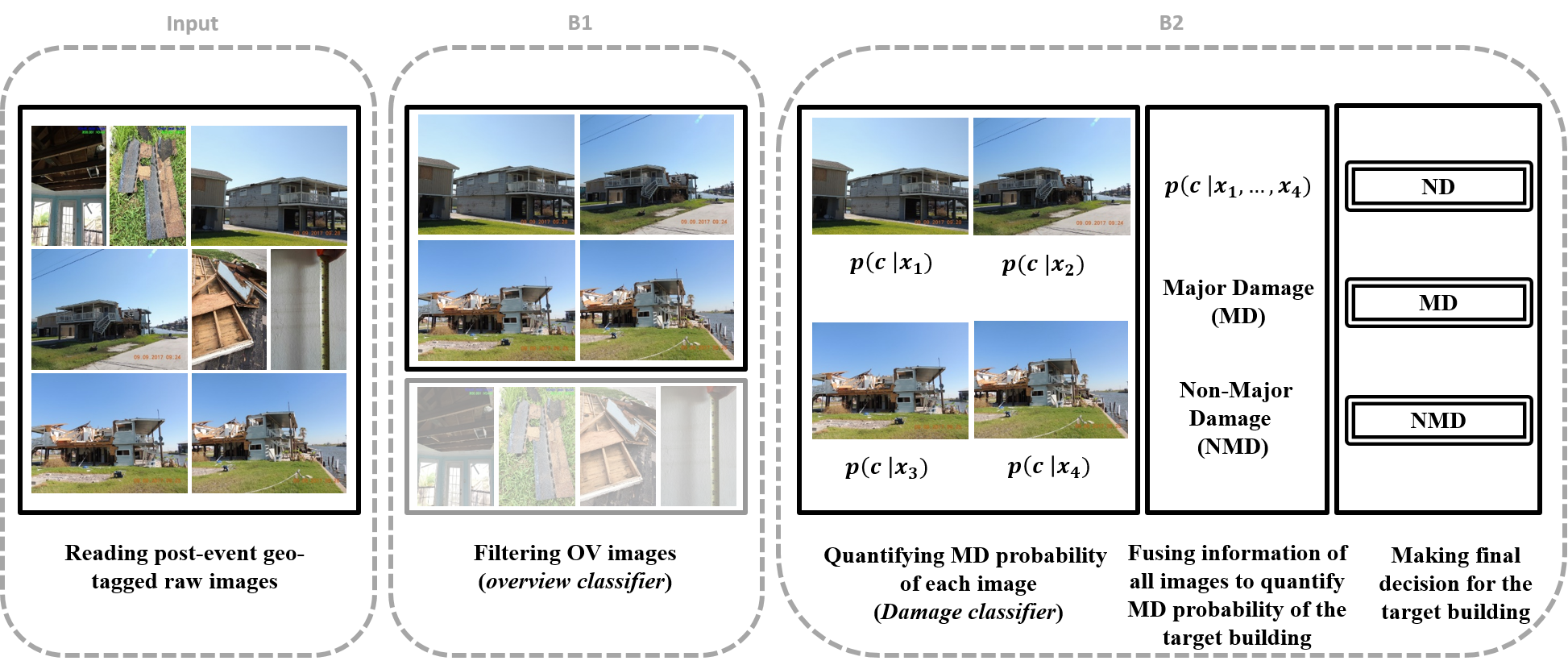}
     \caption{Detailed steps in the post-event data analysis stream.}
     \label{fig:post}
\end{figure*}

The procedure used in the post-event data analysis stream is shown in, Fig. \ref{fig:post}. 
Two classifiers are used for classification of the post-event data, one to filter out less valuable images from the larger set, and a second to determine the condition of the building. 
These are applied to the dataset sequentially, as shown in, Fig. \ref{fig:hir_post}.

The first classifier needed for post-event data analysis is called the \emph{Overview classifier}.
This is a binary classifier that filters flags images that show a sufficient view of the building. 
Each post-event image is classified as either ``Overview'' or ``Non-Overview,'' as indicated in Step \emph{2A}.

\begin{figure*}
	\centering
	\subfloat[OV
	\label{fig:ov}]{\includegraphics[width=0.64\linewidth,height =.5\textheight]{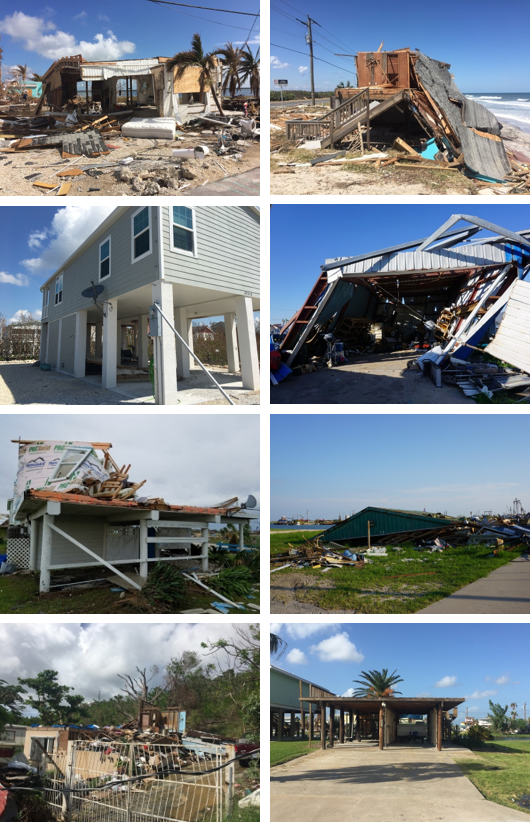}}\hspace*{\fill}
	\subfloat[NOV \label{fig:nov}]{\includegraphics[width=0.32\linewidth,height =.5\textheight]{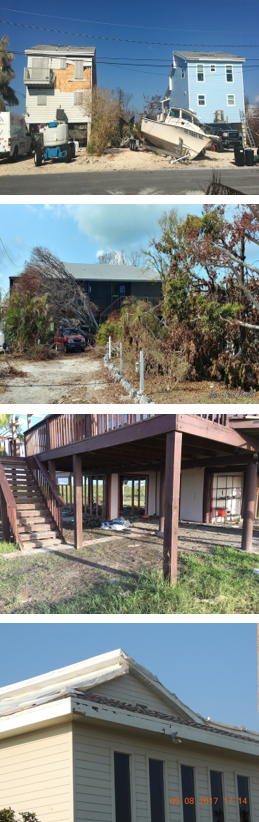}}
	\caption{Samples of images classified as overview (OV) and non-overview (NOV).}
	\label{fig:ovnov}
\end{figure*}

The Overview classifier is defined as:
\begin{itemize}
\item	\textbf{Overview (hereafter, OV):}
Images classified as OV show the entire building, irrespective of whether it is damaged or not, in the sense that they contain more than 70\% of the facade (with either a front view or a side view) and they include portion of the roof.  
To include the possibility of severe damage, an image with some standing columns, or a pile of debris which can clearly be identified as a collapsed building, is also classified as OV.
Examples of the latter include images of the general overall view of standing structural members or a collapsed roof.
An additional restriction of OV images is that no more than 20\% of the image area shows the surrounding buildings.
In some cases, partial obstruction, by trees, cars, and other buildings, is an inevitable challenge. 
However, if the obstruction hides less than 30\% of the building facade, we still consider the image as an OV. 

\item	\textbf{Non-overview (hereafter, NOV):} 
Images that are not OV are NOV.
Examples of NOV include images of the interior of the building, measurements, GPS devices, drawings, multiple buildings, building facades occluded by trees, cars or other buildings.
\end{itemize}
Samples of images defined as OV and NOV are shown in Figs. \ref{fig:ov} and \ref{fig:nov}, respectively.

\begin{figure*}
	\centering
	\subfloat[MD
	\label{fig:MD}]{\includegraphics[width=0.48\linewidth,height =.4\textheight]{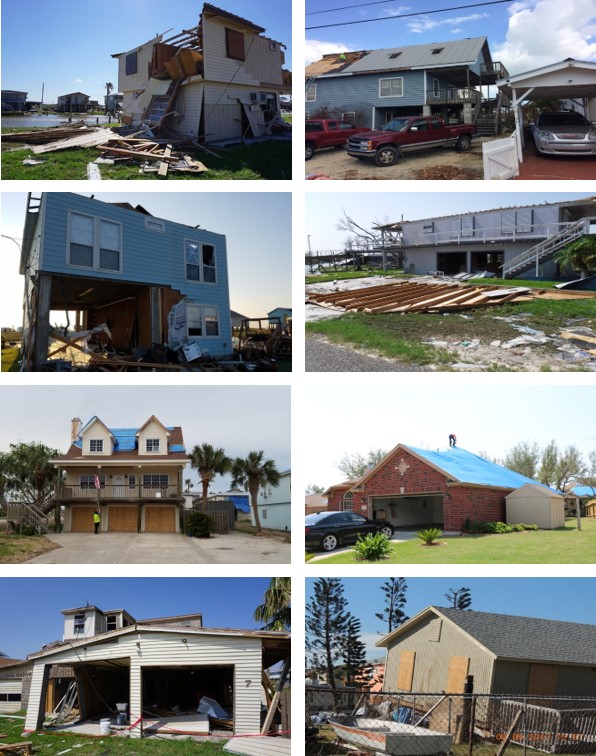}}\hspace*{\fill}
	\subfloat[NMD \label{fig:NMD}]{\includegraphics[width=0.48\linewidth,height =.4\textheight]{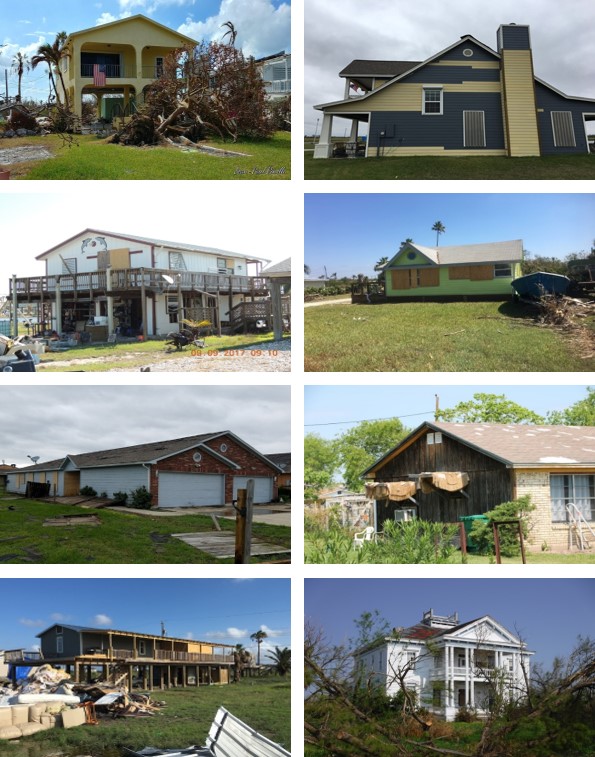}}
	\caption{Samples of images classified as major damage (MD) and non-major damage (NMD).}
	\label{fig:MDNMD}
\end{figure*}

Next, as shown in, Fig. \ref{fig:post}, the subset of images classified as OV are analyzed collectively to determine the overall building condition.
A classifier is trained to determine whether a single OV image should be labeled as ``Major damage'' or ``Non-major damage,'' which includes both minor and no damage.
We call this binary classifier the \emph{Damage classifier}. 
Note that a single image is not sufficient to characterize a building as it may be showing a side from which damage is not visible. 
Therefore, after classifying the damage in each OV image of a given building, the overall condition must be decided by fusing all available information (this will be discussed in Sec. \ref{sec:info_fuse}). 
The Damage classifier is defined as: 
\begin{itemize}
\item	\textbf{Major damage (hereafter, MD):}
Images classified as MD contain visual evidence of severe damaged by wind, wind-driven rain, or flood. 
Specific examples include signs of roof collapse, and column, wall or exterior door failure.
In the case of severe water intrusion/damage, we also classify the image as MD. 
Considerable damage to the roof or exterior doors or windows or garage doors, either from flooding or water intrusion in the case of a hurricane, are also interpreted as major damage. 

\item	\textbf{Non-major damage (hereafter, NMD):}
Images that are not MD are NMD.
No damage, or minor damage, such as cracked, curling, lifted, or missing shingles,  missing flashing, or dents on the doors, are all considered as NMD.

\end{itemize}
Samples of images defined as MD and NMD are shown in Figs. \ref{fig:MD} and \ref{fig:NMD}, respectively.

\subsubsection{Classifiers used in the pre-event stream}
\label{sec:cls_schm_pre}

\begin{figure*}
     \centering
     \includegraphics[width=1\textwidth]{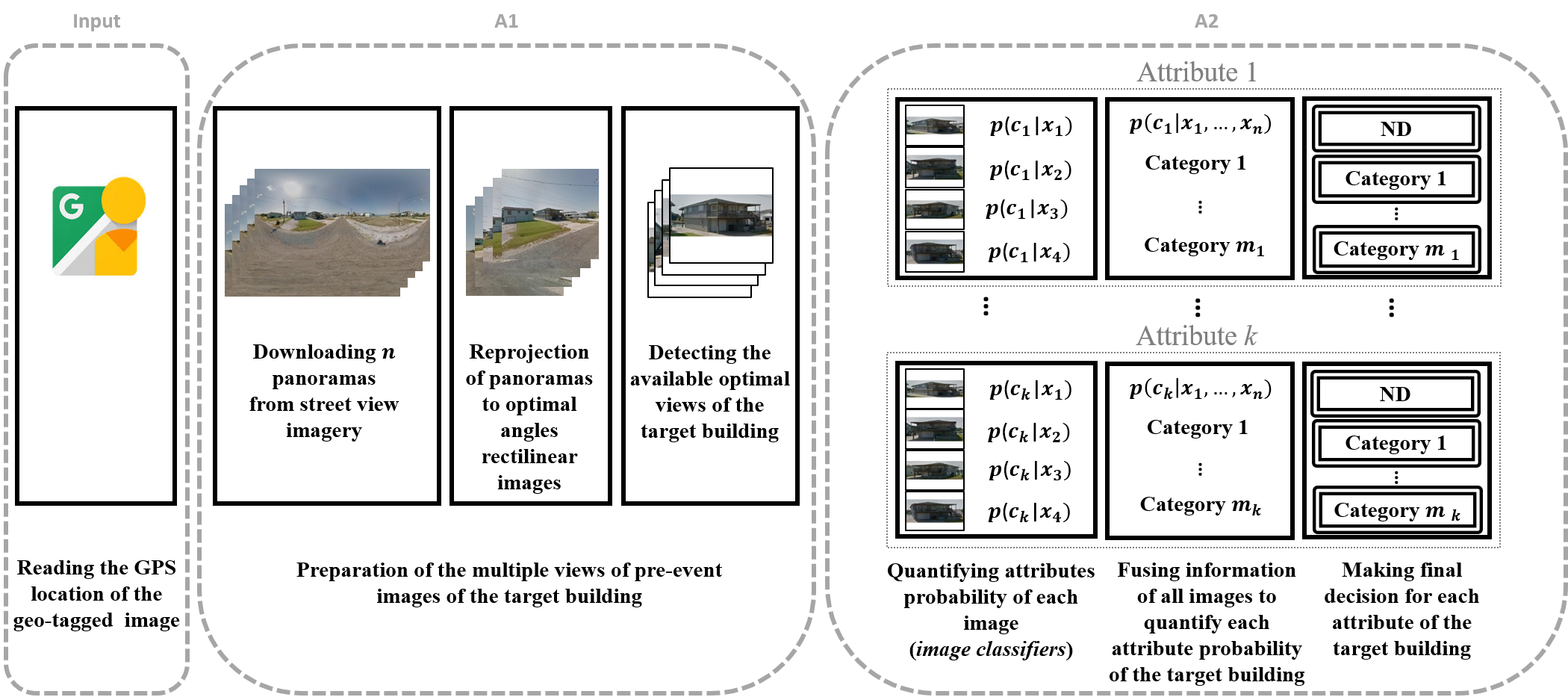}
     \caption{Detailed steps in the pre-event data analysis stream.}
     \label{fig:pre}
\end{figure*}

The sequence of steps used to perform the pre-event data analysis stream is shown in, Fig. \ref{fig:pre}.
In the pre-event stream, multiple external views of each building, collected before the event, are required. 
We employ an automated method we previously developed to extract suitable pre-event residential building images from typical street view panoramas \cite{lenjani2019automated,yeum2018automated}. 

We design three independent classifiers, shown in, Fig. \ref{fig:hir_pre}, to label the scenes containing each view of the pre-event target building.
These classifiers detect: first floor elevation, number of stories, and construction material. 
To successfully train the classifiers to detect building attributes, we need a clear definition of each class. 
In what follows, we describe these definition in detail.

\begin{figure*}
	\centering
	\subfloat[Elevated
	\label{fig:elv}]{\includegraphics[width=0.48\linewidth,height =.5\textheight]{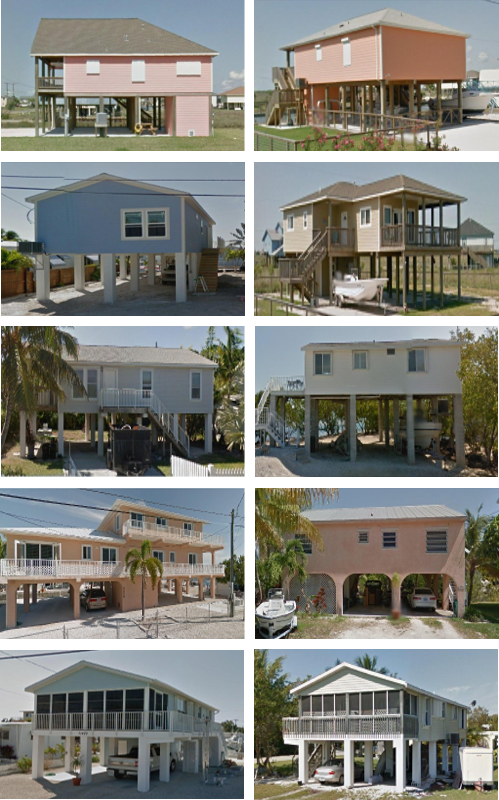}}\hspace*{\fill}
	\subfloat[Non-elevated \label{fig:nelv}]{\includegraphics[width=0.48\linewidth,height =.5\textheight]{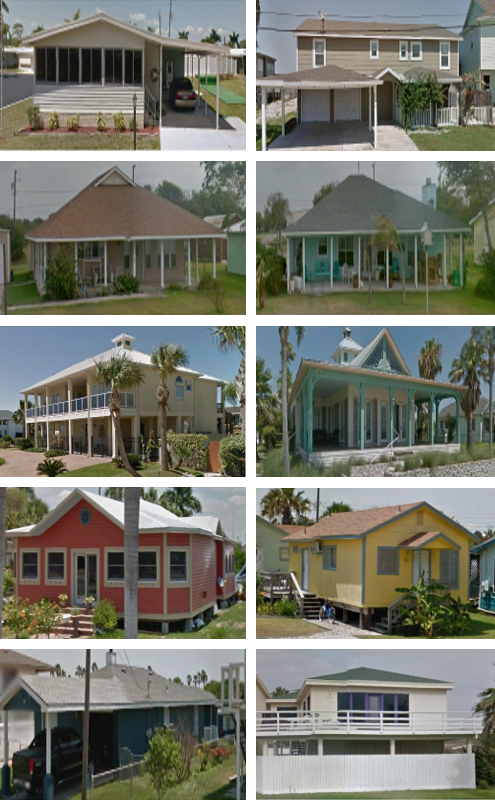}}
	\caption{Samples of images classified as Elevated and Non-elevated building images.}
	\label{fig:elvnelv}
\end{figure*}

One important physical attribute of a residential building is first floor elevation, which is defined as the elevation of the top of the lowest finished floor, which must be an enclosed area, of a building. 
We train a classifier to determine whether a single building image should be classified as ``Elevated'' or ``Non-elevated''.
The \emph{Elevation classifier} is defined as:
\begin{itemize}
\item	\textbf{Elevated (hereafter, EL):}
This class includes buildings with a first floor that appears to be elevated more than 5 feet (or, half a story). 
Buildings are considered as EL when their ground floor, below the first finished floor, is not covered by walls or cladding and is thus visually distinguishable from an occupied floor. 
The lack of coverings or walls is present to potentially allow water to pass through in case of flood to reduce hydrodynamic impact loads. 
In a typical elevated building, the first floor only contains supporting columns (sometimes referred to as slits) which are visually identifiable in the images. Fig \ref{fig:elv} shows samples of EL images.  

\item	\textbf{Non-elevated (hereafter, NEL):}
This class has the opposite meaning as the elevated class. 
It includes images of buildings without first floor elevation, or with a first floor elevation of less than 5 feet. Any images of buildings with a first floor that is covered by walls or cladding are classified as NEL. 
Fig (\ref{fig:nelv}) shows samples of  NEL images.
\end{itemize}

\begin{figure*}
	\centering
	\subfloat[One Story
	\label{fig:1st}]{\includegraphics[width=0.48\linewidth,height =.20\textheight]{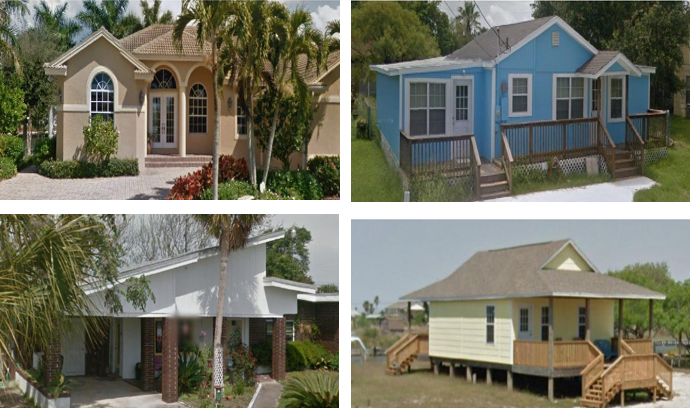}}\hspace*{\fill}
	\subfloat[Two Stories \label{fig:2st}]{\includegraphics[width=0.48\linewidth,height =.20\textheight]{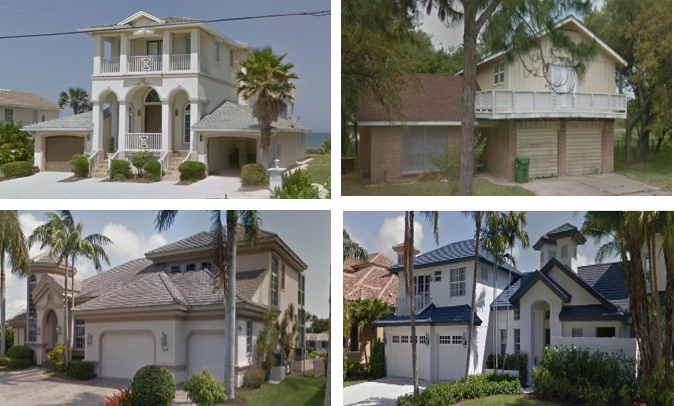}}
	\caption{Samples of images classified as (a) One-story and (b) Two-stories.}
	\label{fig:1and2st}
\end{figure*}

Another useful physical attribute is the number of stories. 
Because we focus on residential buildings here, the vast majority of the images will contain buildings that have either one or two stories.
So, we train a two-class classifier to classify each of the images as either as ``One-story'' or as ``Two-stories.'' 
This classifier does not consider any floors that are not visible, for instance in a case where a floor may be below grade. 
This classifier is the  \emph{Number-of-stories classifier}, and these two classes are defined as follows:
\begin{itemize}
\item	\textbf{One-story (hereafter, 1S):}
This class includes images of buildings which can be considered to have one-story from a structural engineering point of view (i.e., dynamically, it behaves like a single story). 
If any elevation is present in the image, it must not be enough to be classified as EL (i.e., less than about half a story). Fig \ref{fig:1st} shows samples of  \emph{One-story} images. 
\item	\textbf{Two-stories (hereafter, 2S):}
This class includes images of buildings which can be considered to have two-stories, from the structural engineering viewpoint. 
Either a two story building with no first floor elevation, or a one story building with greater than 5 feet of elevation at the first floor is included in the  \emph{Two-stories} category. Fig \ref{fig:2st} shows samples of  \emph{Two-stories} images.
\end{itemize}

\begin{figure*}
	\centering
	\subfloat[Wood
	\label{fig:wood}]{\includegraphics[width=0.24\linewidth,height =.4\textheight]{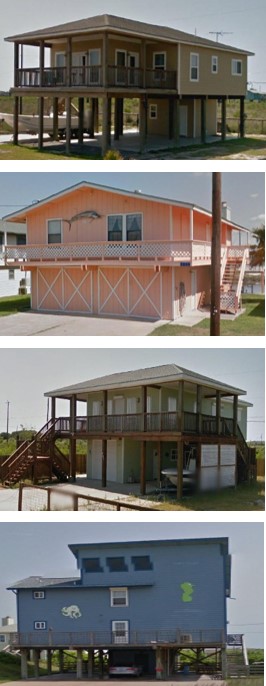}}\hspace*{\fill}
	\subfloat[Masonry \label{fig:masonry}]{\includegraphics[width=0.72\linewidth,height =.4\textheight]{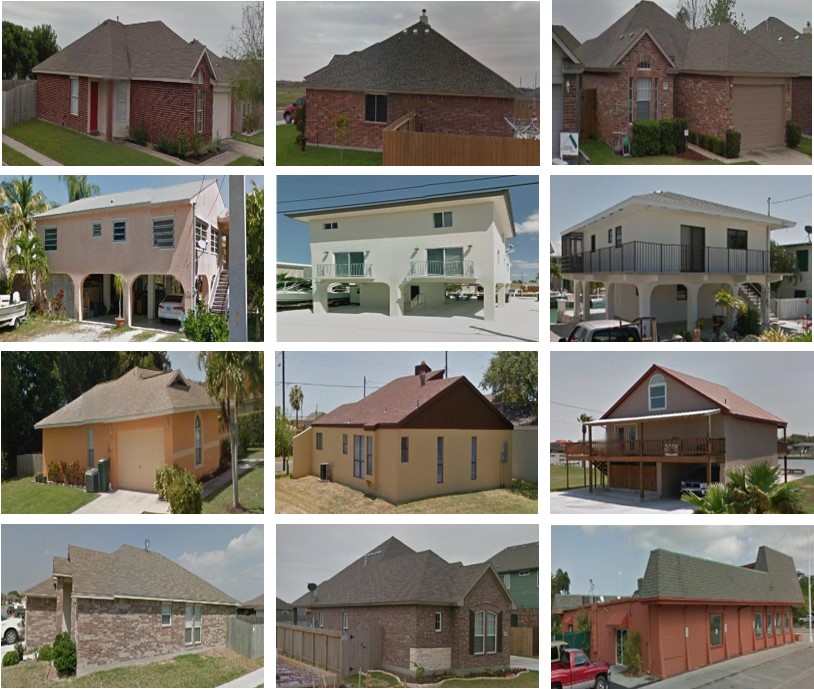}}
	\caption{Samples of images classified as (a) Wood and (b) Masonry.}
	\label{fig:wood_masonry}
\end{figure*}

The third classifier applied to the pre-event images is trained to detect the construction material of the building. In a preliminary survey, it is important to know if wood is the main construction material, or if there is an abundance of other materials present, for instance masonry structural components or veneers. Based on the common construction practices in this geographical region, wood is the main material used for residential construction. The  \emph{Material classifier}, distinguishing between ``Wood'' and ``Masonry,'' is defined as: 

\begin{itemize}
\item	\textbf{Wood (hereafter WO):}
Images in this class provide visible evidence that wood is the main construction material in the building. Note that all materials may not be visible in each image (or even in any image). If all visible parts of the building in the image, including columns, posts, roof structure, exterior load-bearing walls, beams, and girders, are made of wood, the image is classified as WO. Fig \ref{fig:wood} shows samples of WO images.
\item	\textbf{Masonry (hereafter, MA):}
When more than 70\% of the visible portions of the exterior of the building in the image consists of masonry, the image is classified as MA. 
Fig \ref{fig:masonry} shows samples of \emph{Masonry} images. Note that sloped roof buildings with masonry walls generally have wooden roofs.
\end{itemize}

\subsection{Information fusion}
\label{sec:info_fuse}

We discuss how to make decisions using a probabilistic approach that fuses the classification results from several images.
Let $C$ be the random variable (r.v.) corresponding to a given physical building attribute taking values in the set $\mathcal{C}$.
Now consider $n$ images $x_1,\dots,x_n$ of the same building and let $C_1,\dots,C_n$ be the set of r.v.'s corresponding to the detection of the physical attribute each one of the images.
The $C_i$'s also take values in $\mathcal{C}$, but they are distinctly different.
The former, $C_i$, only tells us which attribute was detected in image $i$, whereas the latter, $C$, which attribute was detected in the entire building.
The two are different because an attribute may not be visible in all images.
Since $C_i$ depends only on the $i$-th image, we have:
\begin{equation}
    \label{eq:1}
    p(C_i = c_i | x_1,\dots, x_n) = p(C_i = c_i | x_i) := f_{\text{CNN},c_i}(x_i),
\end{equation}
where $f_{\text{CNN},c}(x)$ is the CNN-based classifier corresponding to the attribute.
How can we use the classification of each image ($C_i$) to classify the entire building ($C$)?
We have:

\begin{strip}
\begin{eqnarray}
\label{eq:2}
\begin{array}{ccc}
p(C=c|x_1,\dots,x_n)
&=&\sum_{c_1,\dots,c_n\in\mathcal{C}} p(C=c|C_i=c_1,\dots,C_n=c_n, x_1,\dots,x_n)\cdot
p(C_1=c_1,\dots,C_n=c_n|x_1,\dots,x_n)\\ 
&=&\sum_{c_1,\dots,c_n\in\mathcal{C}} p(C=c|C_i=c_1,\dots,C_n=c_n)\cdot
p(C_1=c_1,\dots,C_n=c_n|x_1,\dots,x_n)\\ 
&=&\sum_{c_1,\dots,c_n\in\mathcal{C}}p(C=c|C_i=c_1,\dots,C_n=c_n)\cdot
\prod_{i=1}^np(C_i=c_i|x_1,\dots,x_n)\\ 
&=&\sum_{c_1,\dots,c_n\in\mathcal{C}}p(C=c|C_i=c_1,\dots,C_n=c_n)\cdot
\prod_{i=1}^np(C_i=c_i|x_i).
\end{array}
\end{eqnarray}
\end{strip}

Here, going from the first to the second step we assumed that the raw data $x_1,\dots,x_n$ do not provide any additional information about the building label $C$ if image labels $C_1,\dots,C_n$ are known.
This assumption is discussed again in Sec. \ref{sec:post_info_fuse}.
For the next steps, we use the sum rule of probability, and observe that the $C_i$'s are independent conditional on the images, and then apply Eq.~(\ref{eq:1}), w
The term $p(C=c|C_1=c_1,\dots,C_n=c_n)$ gives the probability that the target building is labeled $c$, given the available images are labeled as $c_1,\dots,c_n$.
This \emph{fusion} probability is attribute-specific, as discussed in Secs. \ref{sec:post_info_fuse} and \ref{sec:pre_info_fuse} for post-event and pre-event attributes, respectively.
Note that, in our case, the set of possible classes $\mathcal{C}$ always contains two elements.
Without loss of generality, in what follows, we are going to denote it with $\mathcal{C} = \{0, 1\}$ with $c=1$ corresponding to the positive detection of an attribute and $c=0$ to detection of the alternative.

Finally, let $\mathcal{D}$ be the set of possible decisions that are available to us with regard to a given building, and one void class, here called No Decision (ND), added to skip making a decision when a confident decision is not available.  
For example, in case of predicting the overall damage condition, it will include MD, NMD and ND.
Define a loss function denoted $\ell(d,c)$ which represents the resulting loss if we choose decision $d$ in $\mathcal{D}$ when the true attribute is $c$ in $\mathcal{C}$.
Ignoring risk preferences, the rational decision is the one minimizing the expected loss:
\begin{equation}
\label{eq:3}
    d^*(x_1,\dots,x_n) = \arg\min_{d\in\mathcal{D}} \sum_{c\in\mathcal{C}}\ell(d,c)p(C=c|x_1,\dots,x_n).
\end{equation}
Here, the loss represents the threshold for making a decision about the building or leaving it as ND. 
The loss function parameters can be tuned by the reconnaissance teams for a specific reconnaissance goal, such as to either make the best possible decision about all cases, or to make decision only when it is highly confident.
The loss function is structured to handle the trade-off between the accuracy and informativeness of the results through adding ND class to skip making a decision in case of not being sufficiently confident.

\subsubsection{Post-event}
\label{sec:post_info_fuse}
The case of the post-event stream, and in particular the MD ($C=1$) vs NMD ($C=0$) problem, is inherently asymmetric.
On one hand, one must consider the whether or not the set of images shows the building from all sides.
For example, a single image classified as NMD is not sufficient to conclude that the building is indeed NMD since the damage may simply not be visible from the viewpoint of that image.
So, to classify a given building as NMD, we need to ensure that all sides of the building are shown in the set of images (in this case, we say that \emph{the building is covered}).
If all of these individual images are classified as NMD, only then can the building be categorized as NMD.
On the other hand, to classify a building as MD, it is sufficient to have a single image classified as MD.

Define a binary r.v. $Z$ taking values $\{0,1\}$ indicating that the building is not covered and is covered, respectively.
Let $p(Z=1 | C_1=c_1,\dots, C_n=c_n, x_1, \dots, x_n)$ be probability that the available images sufficiently cover the target building, hereafter \emph{coverage probability}.
Our dataset does not provide any information about $Z$ (the images do not include sufficient geolocation information).
Therefore, we may write:
\begin{equation}
\label{eq:5}
\begin{split}
p(Z=1 | C_1=c_1,\dots, C_n=c_n, x_1, \dots, x_n) & = \\
p(Z=1 | C_1=c_1,\dots, C_n=c_n) & = q_n,
\end{split}
\end{equation}
where in the last step we used the observation that only the number of images are affects our state of knowledge about $Z$, i.e., the labels themselves are uninformative about Z.
Obviously, $q_1 = 0$ and $q_2 = 0$ since one or two images cannot cover the building.
Furthermore, we should have that $0\le q_i \le q_{i+1}\le 1$.
The specific numerical choice of this series of probabilities depends on our state of knowledge about how the data were collected.
For example, if we knew that any three images cover the building, then we would set $q_1 = q_2 = 0$ and $q_n = 1$ for $n\ge 3$.

Now, we use the sum rule on the fusion probability:

\begin{strip}
\begin{equation}
\label{eq:6}
\begin{split}
p(C=1|C_1=c_1,\dots,C_n=c_n) & = p(C=c|C_1=c_1\dots,C_n=c_n, Z=1)  
p(Z=1 | C_1=c_1,\dots, C_n=c_n) \\ &
+ p(C=1|C_1=c_1,\dots,C_n=c_n, Z=0)  
p(Z=0 | C_1=c_1, \dots, C_n=c_n)\\
& = p(C=1|C_1=c_1\dots,C_n=c_n, Z=1)  
q_n \\ &
+ p(C=1|C_1=c_1,\dots,C_n=c_n, Z=0)  
(1-q_n).
\end{split}
\end{equation}
\end{strip}

The two terms that we need to specify are the probabilities of labeling the building as MD ($C=1$) given the image labels and whether or not the building is covered.
For the covered case, we set:
\begin{equation}
\label{eq:7}
p(C=1|C_1=c_1\dots,C_n=c_n, Z=1) = \ceil*{\frac{ \sum_{i=1}^{n} c_i} {n}},
\end{equation}
where $\ceil{\cdot}$ is the first integer greater than its argument.
This means that there is at least one image labeled as MD, then the entire building is labeled MD.
For a covered building to be labeled NMD, all images must be labeled NMD.
There are no intermediate cases.
For the uncovered case, we set:
\begin{equation}
\label{eq:8}
\begin{split}
p(C=1|C_1=c_1,\dots,C_n=c_n, Z=0) & = \\ \max \left\{ \ceil*{\frac{ \sum_{i=1}^{n} c_i} {n}} , \theta_n \right\},
\end{split}
\end{equation}
where $\theta_n$ represents the probability that the building is MD but the damage is not visible in $n$ images.
Again, $\theta_n$ depends on what we know about data collection.
In general, we must have $0\le \theta_i \le \theta_{i+1}\le 1$.
In our case studies, we simply pick $\theta_n=0.5$ for all $n$.
So, for the uncovered case, a single MD labeled image is sufficient to characterize the building as MD.
However, if all images are labeled NMD, there is still a probability, $\theta_n$, that the building is MD but the damage is not visible.

\subsubsection{Pre-event}
\label{sec:pre_info_fuse}
In the pre-event stream, we detect binary physical attributes, i.e., EL vs NEL, 1S vs 2S, and WO vs MA.
All these cases are similar in nature.
The more often an attribute is detected in the images the more likely it is really there.
The simplest model that encodes this intuition is:
\begin{equation}
\label{eq:4}
p(C=1|C_1={c}_1,\dots,C_n={c}_n) = \frac{ \sum_{i=1}^{n} c_i}{n}.
\end{equation}
Here, we exploit the 0-1 encoding of the binary class.
The probability on the right hand side is simply the average number of ones in the $n$ images.
Essentially, the r.v. $C$ conditional on the r.v.'s $C_1,\dots,C_n$ has a Bernoulli distribution.
The approach can be trivially generalized, using a Categorical distribution, to the case where $\mathcal{C}$ contains more than two options.

\section{Experimental validation}
\label{sec:exp_val}

We verify the individual classifiers and validate the overall technique using a high-quality published and curated post-event dataset.  
These perishable information were captured during reconnaissance missions that took place shortly after hurricanes Harvey and Irma, led by the NSF-funded Structural Extreme Events Reconnaissance (StEER) Network, with data collection supported by the Fulcrum App \cite{steer_data}.
We have tried three networks, Inception v3 \cite{szegedy2016rethinking}, InceptionResNetV2 \cite{szegedy2017inception}, and Xception \cite{chollet2017xception},  as the image classifiers, and Xception network slightly outperformed the two others.
We implemented Xception with Depthwise Separable Convolutions network, in Keras \cite{chollet2015keras}. 

In this implementation we used Stochastic Gradient Descent(SGD) optimizer. 
The SGD hyper-parameters used for the classifiers were fine-tuned using grid search to train each of the classifiers.
We tuned the hyper-parameters, particularly the learning rate which is the most important hyper-parameter \cite{goodfellow2016deep}, carefully to improve the performance of the classifiers.
We set the grid to search for 1) learning rate in ~$\{\num{1e-1}, \num{5e-2}, \dots, \num{5e-9}, \num{1e-10}\}$ 2) momentum in ~$\{\num{1e-1}, \dots, \num{9e-1} \text{ and } \num{99e-2}\}$ 3) weight decay coefficient in ~ $\{\num{1e-1}, \dots, \num{1e-10}\}$.
We randomly separate the train and test set with $70\%$ and $30\%$ , respectively, of the data for each classifier. 
To avoid over-fitting, we randomly sample out $10\%$ of the train set to use for hyper\--para\-meters fine-tuning. 
Table \ref{tab:hyperpar}, shows the hyper-parameters used to train these five required classifiers.

\begin{table*}
\caption{Hyper-parameters used to train the classifiers}
\label{tab:hyperpar}
\centering
\begin{tabular}{l l l l}
\toprule
\textbf{Classifier name} \hspace{40pt} & \textbf{Initial learning rate} & \textbf{Momentum} & \textbf{Weight decay coefficient} \\
\cmidrule{1-1}
\cmidrule{2-4}
Overview & $\num{1e-5}$  & $\num{9e-1}$ & $\num{1e-6}$\\
Damage & $\num{5e-6}$ & $\num{9e-1}$ & $\num{1e-6}$\\
Elevation & $\num{1e-6}$  & $\num{9e-1}$ & $\num{1e-6}$\\
Number-of-stories & $\num{5e-7}$ & $\num{9e-1}$ & $\num{1e-7}$\\
Material & $\num{1e-7}$  & $\num{9e-1}$ & $\num{1e-7}$\\
\bottomrule
\end{tabular}
\end{table*}

\begin{figure*}
     \centering
     \includegraphics[width=1\textwidth]{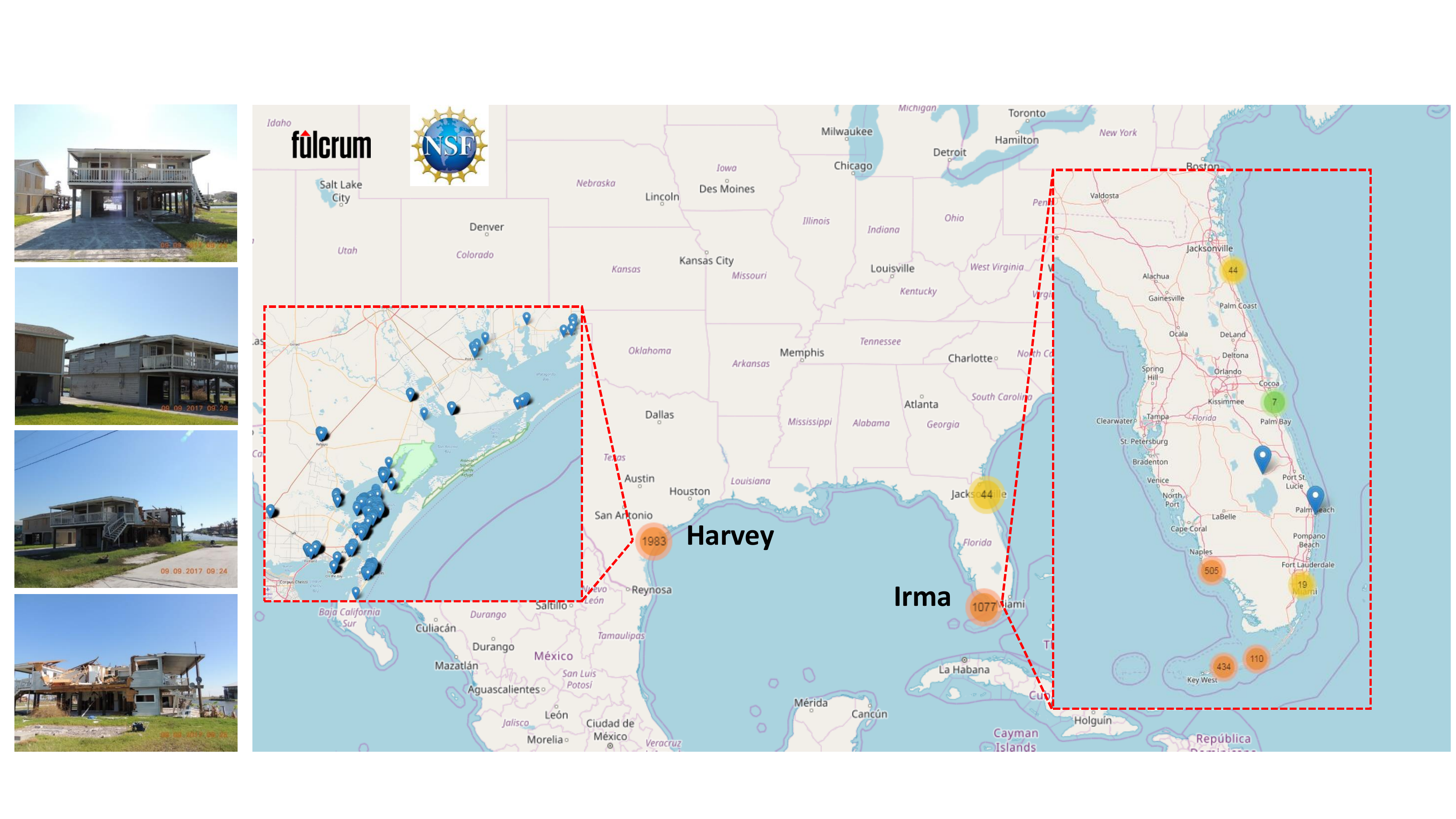}
     \caption{Post-event reconnaissance dataset collected after Harvey and Irma and published on DesignSafe-CI and Fulcrum \cite{designsafe,fulcrum,steer_data}.}
     \label{fig:data}
\end{figure*}

The StEER network was formed to document the damage induced and enable research to understand the effects of a series natural hazard events \cite{nsfgrantsteer,steer}, including hurricanes Harvey, Irma and Maria in 2017 \cite{kijewski2018hurricane_harv,pinelli2018overview_irm}, and hurricane Florence and Michael in 2018 \cite{kijewski2018steer_flor,kijewski2018steer_michfat,kijewski2018steer_michvat}, on the built environment. 
An overview of the dataset \cite{steer,fulcrum} is shown in, Fig. \ref{fig:data}. 
Detailed damage surveys of more than 4,000 buildings were conducted door-to-door \cite{steer_data, designsafe}. 
The data include assessments of the post-event condition of most of the buildings. 
Other documentation includes primary structural typologies, construction materials, and certain component damage levels. 
The documentation available for this data also includes both building attributes plus observations of the overall damage condition of the building after the hurricane. 
Thus, these data are well-suited for validation of the technique developed. 

For training the classifiers we used data from 3,141 buildings, including 2,020 buildings collected after hurricane Harvey in Texas, and 1,121 building collected after hurricane Irma in Florida. 
The data vary greatly from building to building in terms of completeness and number of images collected. 
Thus, not all the data collected from these 3,141 buildings are useful. 
We pre-process the dataset as follows. 
We made adjustments to the pre-event attributes documented in the original dataset that were necessary to conform with our definitions. 
The first floor elevation is reported as an estimated height of elevation in the original documentation. 
Here, we use our threshold of 5 feet to manually label the data for training, testing and validation.
Then, if the building is elevated, we also add one to the number of stories reported to conform to our definition.
Regarding the construction material, we make use of the attribute in the original data called structural framing. 
However, most of these building actually use wood for the structural framing, or the load bearing elements, and thus we redefine it as the main construction materials visible on the exterior of each building as explained in Sec. \ref{sec:cls_schm_pre}. 
When multiple items are provided in the original data, we simply use the first material listed. 

Because the data we use for validation do not contain geo-location information, we only consider the number of available images (see Sec. \ref{sec:post_info_fuse}). 
In Sec. \ref{sec:post_info_fuse}, we defined the probability that $n$ images are sufficient to cover the building as $p(Z=1 | C_1=c_1,\dots, C_n=c_n) = q_n$.
Currently the typical number of images captured in wind-event reconnaissance missions is quite small. 
Furthermore, there is a certain bias in the collection process since the data collector is, typically, interested in collecting images of damage.
For example, we observe that data collectors take fewer images of buildings that have no damage or only minor damage. 
In these circumstances, if only one image is captured, then we may conclude that the building is sufficiently covered, i.e.,  $q_n=1$ for all $n\ge 1$. 
In a more objective data collection process, one has to adjust coverage probability accordingly, see Sec. \ref{sec:mehtod_val_post}.

\begin{figure*}
	\centering
	\subfloat[Accuracy of the classifiers for post-event stream.
	\label{fig:post-acc}]{\includegraphics[scale=0.5]{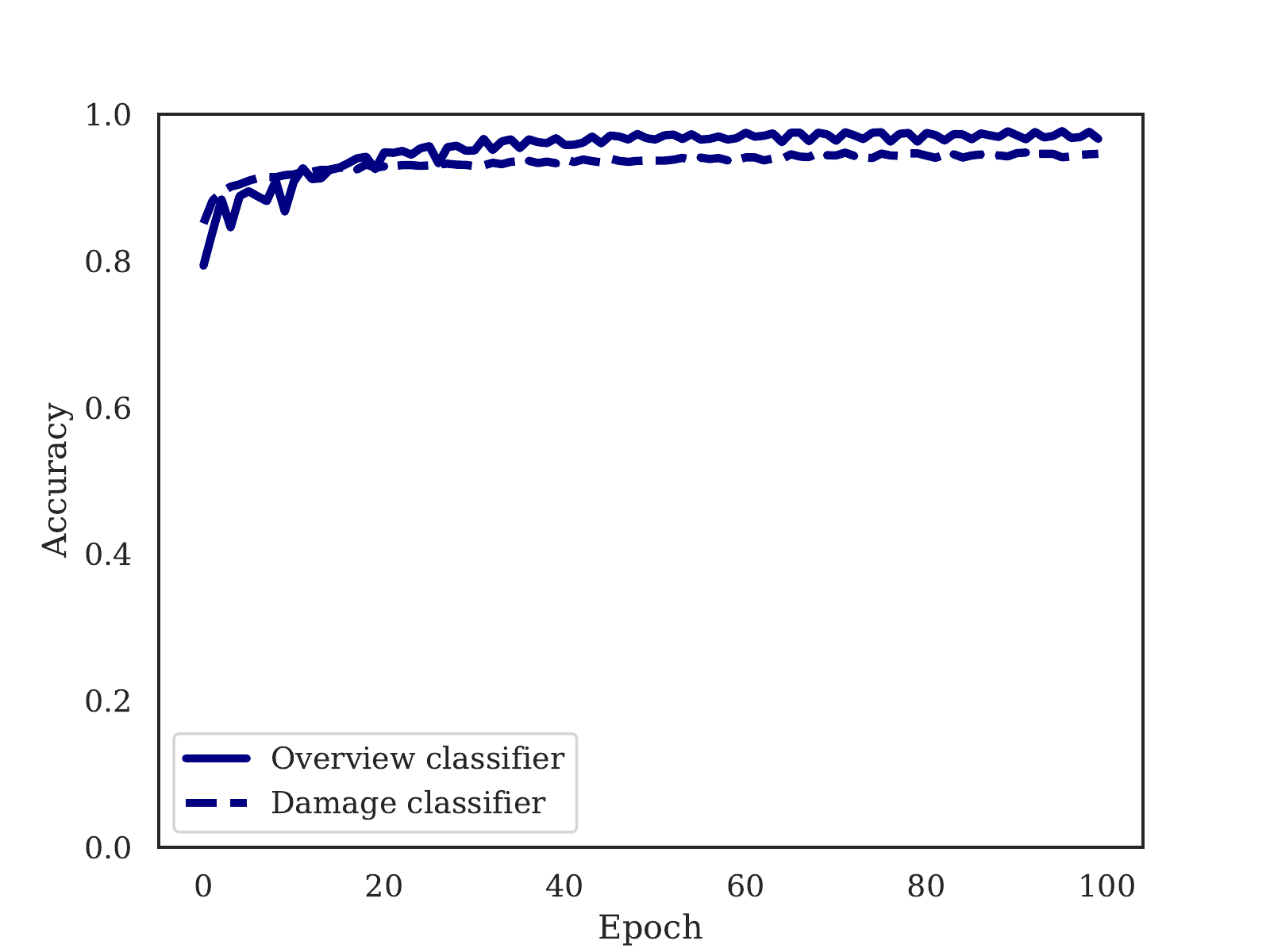}}
	\subfloat[Accuracy of the classifiers for pre-event stream. \label{fig:pre_acc}]{\includegraphics[scale=0.5]{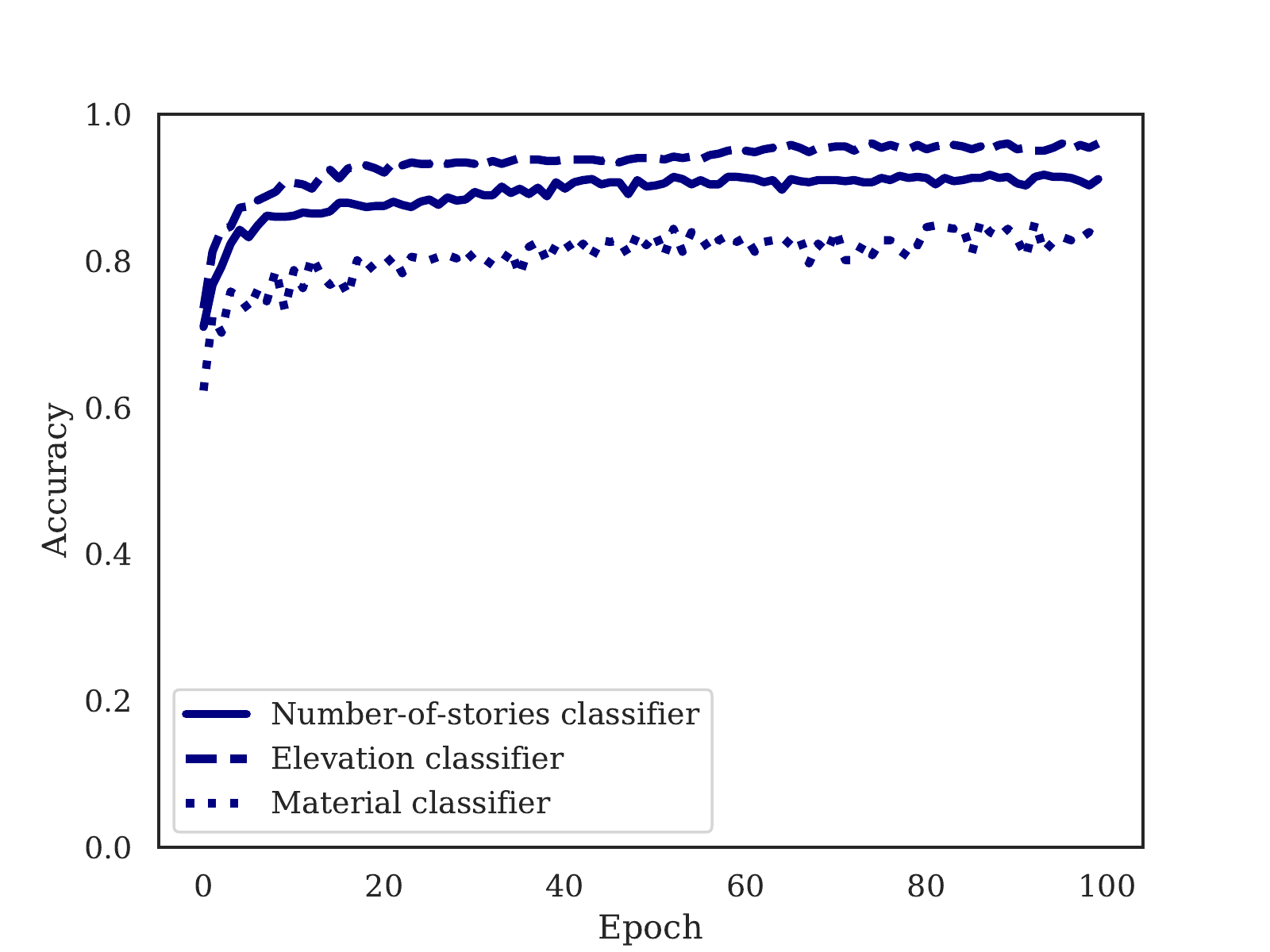}}
	\caption{Accuracy plots of classifiers.}
	\label{fig:acc}
\end{figure*}

We evaluate the performance of the pre-event and post-event data analysis streams independently. 
The validation of the method involves first evaluating the performance of the individual steps in each branch (i.e., of each classifier), as well as considering the end-to-end performance of each data analysis branch. 
Fig.\ref{fig:post-acc} and \ref{fig:pre_acc} show only the accuracy of the classifiers used for post-event and pre-event stream, respectively. However we evaluate the end-to-end performance of the method developed in Sec. \ref{sec:mehtod_val_post} and Sec. \ref{sec:mehtod_val_pre}.
The input to each branch is the set of geo-tagged raw images of the buildings. 
To validate each of these, we use raw available data from all of the 1,121 buildings collected after hurricane Irma. 
Here we explain both the post-event and pre-event data analysis streams validation results.
In the post-event stream, first we demonstrate the results for an example loss function assuming all buildings are sufficiently covered.
Then, we discuss how the results can be improved if we refine the coverage probability, $q_n$ in Eq. \ref{eq:5}.
Subsequently, we study the effect of the loss function parameters on the trade-off between \emph{accuracy} and \emph{ND rate}, rate of ND predictions over all permissible predictions.
In the post-event stream, we illustrate the results for an example loss function, and then the procedure for tuning of the loss function parameters is discussed.

\subsection{Post-event stream validation}
\label{sec:mehtod_val_post}

As described earlier, each OV post-event image is passed through the damage classifier.
Predicting the overall condition of the building, based only on images, is subject to error, see Sec. \ref{sec:post_info_fuse}. 
Even if the building is covered, it may still be difficult to make the decision based entirely on the images.
For example, the damage shown in the image may not be sufficiently severe to be labeled MD, nor minor enough to confidently labeled as NMD.
Under these circumstances, even human inspectors face difficulties and the situation calls for a more detailed inspection.

\begin{table}
\caption{Loss function.}
\label{tab:post-loss_func}
\centering
\begin{tabular}{c c c c c c}
&       &\multicolumn{3}{c}{Decision}\\
&      &  \cellcolor{gray!10}ND& \cellcolor{gray!10}MD & \cellcolor{gray!10}NMD \\
&\cellcolor{gray!10}MD & $\alpha_1$  & 0 & 1 \\
&\cellcolor{gray!10}NMD & $\alpha_2$ & 1 & 0\\
\rowcolor{gray!10} \cellcolor{white}
\rot{\rlap{\hspace{10pt}True label}}
\end{tabular}
\end{table}

The general form of the loss function is shown in, Table~\ref{tab:post-loss_func}.
Without loss of generality, we can set the loss of correct predictions to zero.
The cost of mistakenly characterizing an MD (NMD) building as NMD (MD) is 1.
The cost of labeling as ND when the building state is MD (NMD) is $\alpha_1$ ($\alpha_2$). 
These parameters are selected to reflect the  goals of the preliminary survey, see Sec. \ref{sec:tunlos}. 

\subsubsection{Sample results}
\label{sec:sampres}

\begin{figure}
     \centering
     \includegraphics[scale=.5]{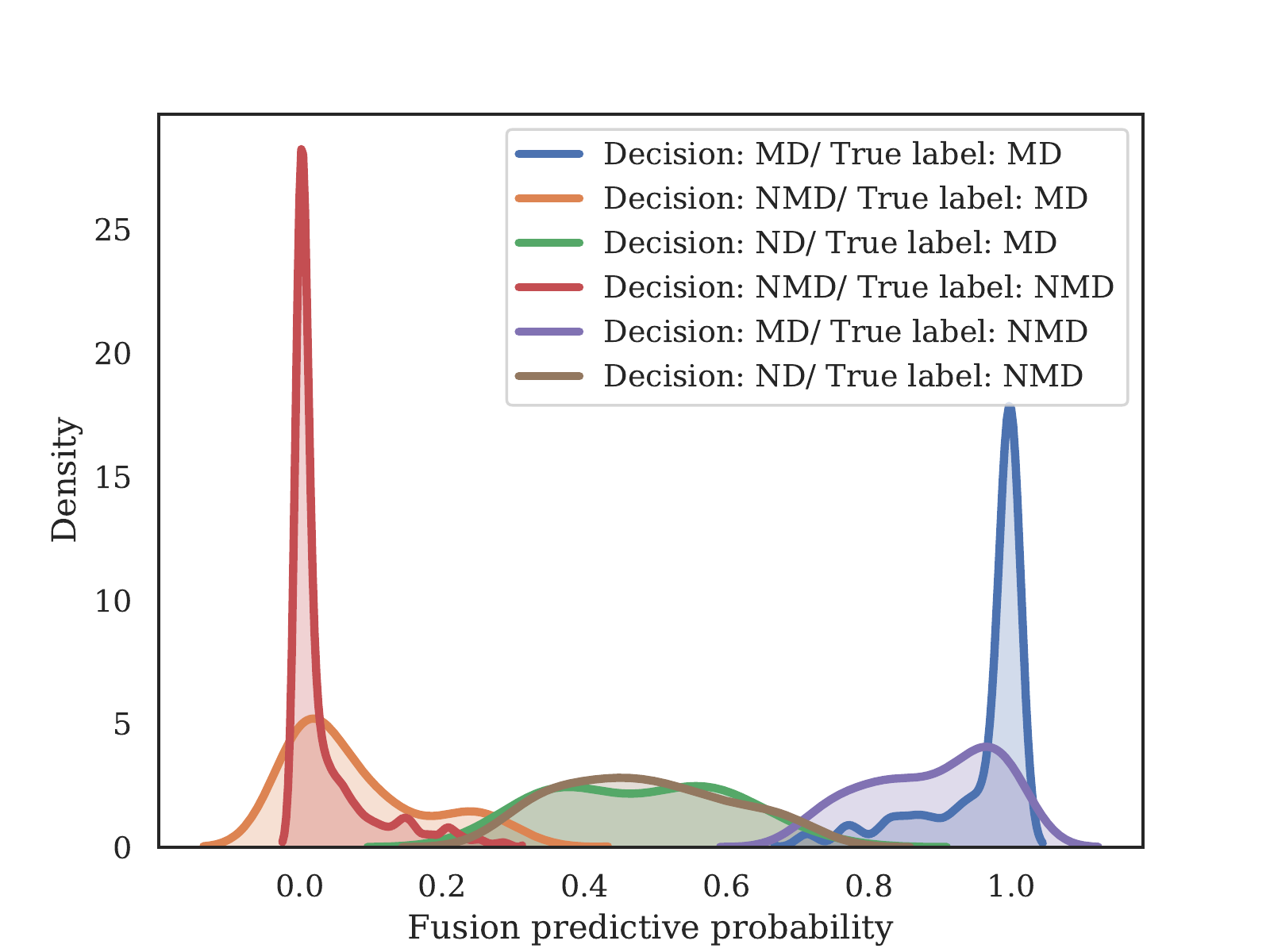}
     \caption{Density of the fusion predictive probabilities corresponding to each different decision, assuming  all  buildings are  sufficiently covered (with $q_n =1$ for $ n \geq 1 $).}
     \label{fig:fuspr}
\end{figure}

\begin{table}
\caption{Confusion matrix using a loss function with parameters ($\alpha_1 = \alpha_2 = 0.3$) ; assuming all buildings are sufficiently covered.}
\label{tab:conf_setup1}
\centering
\begin{tabular}{c c c c c c c}
&       &\multicolumn{3}{c}{Decision}\\
&      &  \cellcolor{gray!10}No OV & \cellcolor{gray!10}ND& \cellcolor{gray!10}MD & \cellcolor{gray!10}NMD & \cellcolor{gray!50}All\\
&\cellcolor{gray!10}No label& 26 & 6 & 5 & 17 & 54\\
&\cellcolor{gray!10}MD & 44  & \cellcolor{yellow!20}16 & \cellcolor{green!20}151 & \cellcolor{red!20}39 & 250\\
&\cellcolor{gray!10}NMD & 109 & \cellcolor{yellow!20}71 & \cellcolor{red!20}71 & \cellcolor{green!20}566 & 817\\
&\cellcolor{gray!50}All & 179 & 93 & 227 & 622 & 1,121\\
\rowcolor{gray!10} \cellcolor{white}
\rot{\rlap{\hspace{10pt}True label}}
\end{tabular}
\end{table}
First, consider the case in which all of the buildings are assumed to be captured adequately with the images available, $q_n =1$ for all $n\ge 1$, and pick a loss function with $\alpha_1 = \alpha_2 = 0.3$.
This choice of the loss function making mistakes has a unit cost, while not deciding costs thirty percent of the mistake cost.
In Fig. \ref{fig:fuspr}, we visualize the density of the fusion predictive probabilities corresponding to each different decision and true label, i.e., density of decisions made at a given fusion probability.
It shows six combination of the two true labels, MD and NMD, and three possible decisions, MD, NMD and ND. 
The correct decisions for the buildings with NMD (MD) true labels, depicted in red (blue), show low-variance right (left)-skewed density with a mode close to 0 (1).
However, the densities of the incorrect decisions for both MD and NMD buildings, have more variance. 
Table \ref{tab:conf_setup1} provides the confusion matrix, table of true labels versus predicted, for the results of our demonstration of the end-to-end post-event stream data analysis. 
Out of a total of 1,121 buildings visited after hurricane Irma, the dataset includes 54 buildings with no true label, and 179 buildings with no OV images. 
Also, 26 buildings are not distinct and those data are merged into one building set. 
Therefore we have 914 labeled buildings with OV images.
The results show that 717 buildings are correctly categorized,  110 buildings are classified incorrectly, and 87 buildings labeled ND. 

\begin{figure*}
	\centering
    \includegraphics[width=1\textwidth]{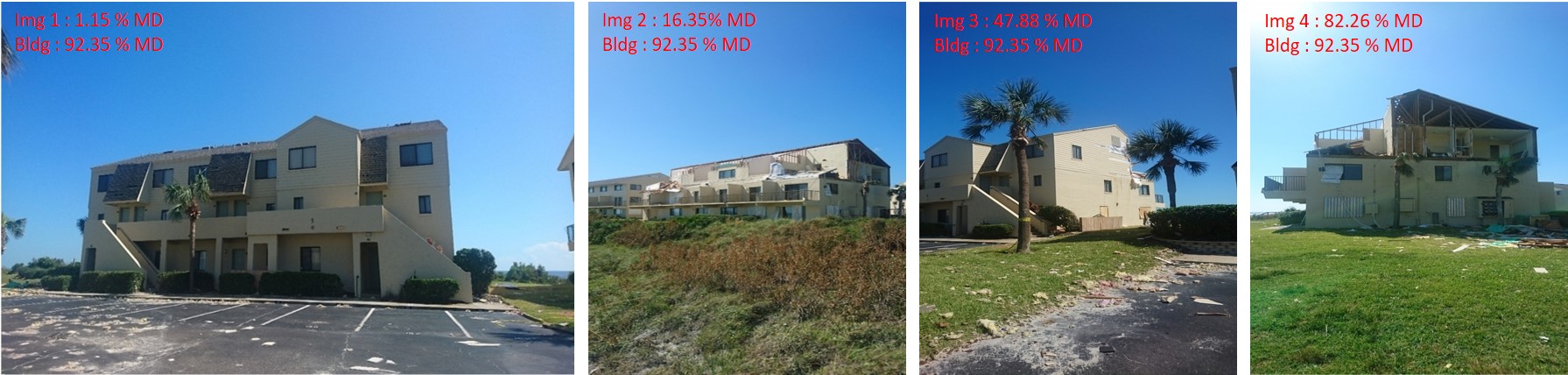}
    \caption{Sample of a correct MD detection.}
	\label{fig:sample_res_MD}
\end{figure*}

\begin{figure*}
	\centering
    \includegraphics[width=1\textwidth]{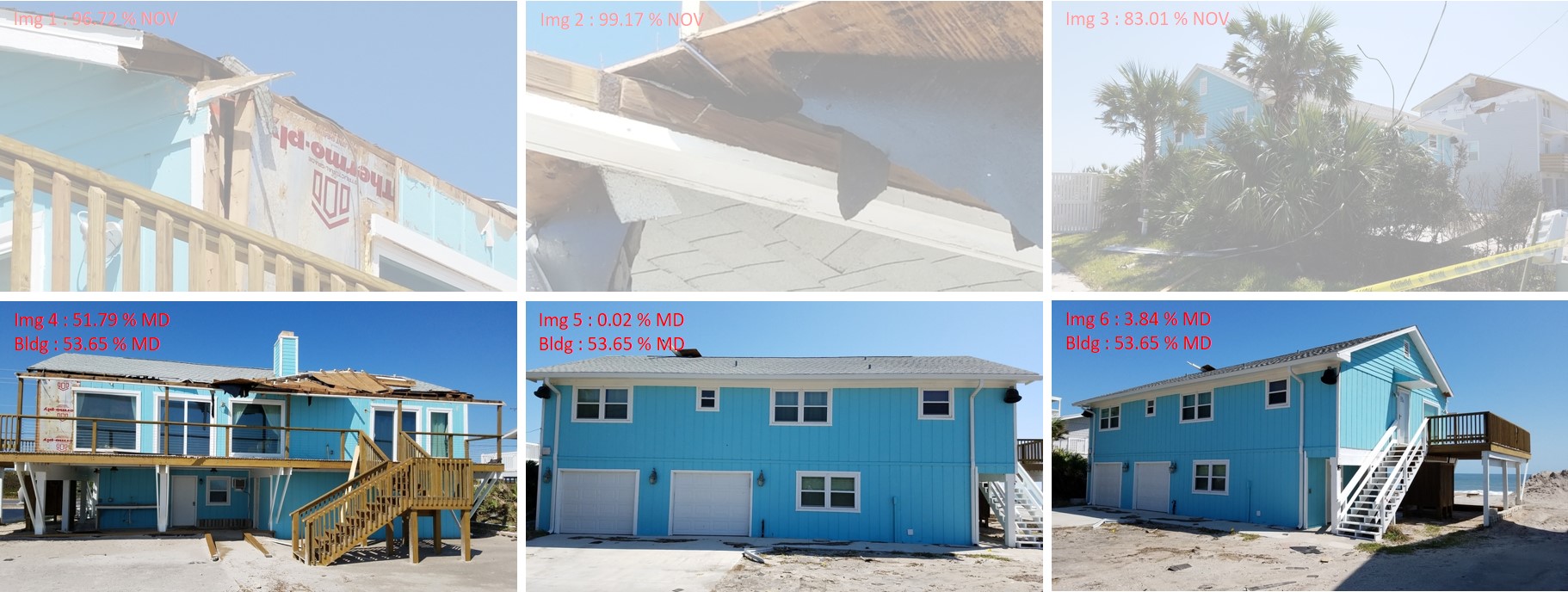}
    \caption{Sample of a ND building categorization.}
	\label{fig:sample_no_dec}
\end{figure*}

\begin{figure*}
	\centering
    \includegraphics[width=1\textwidth,height =.2\textheight]{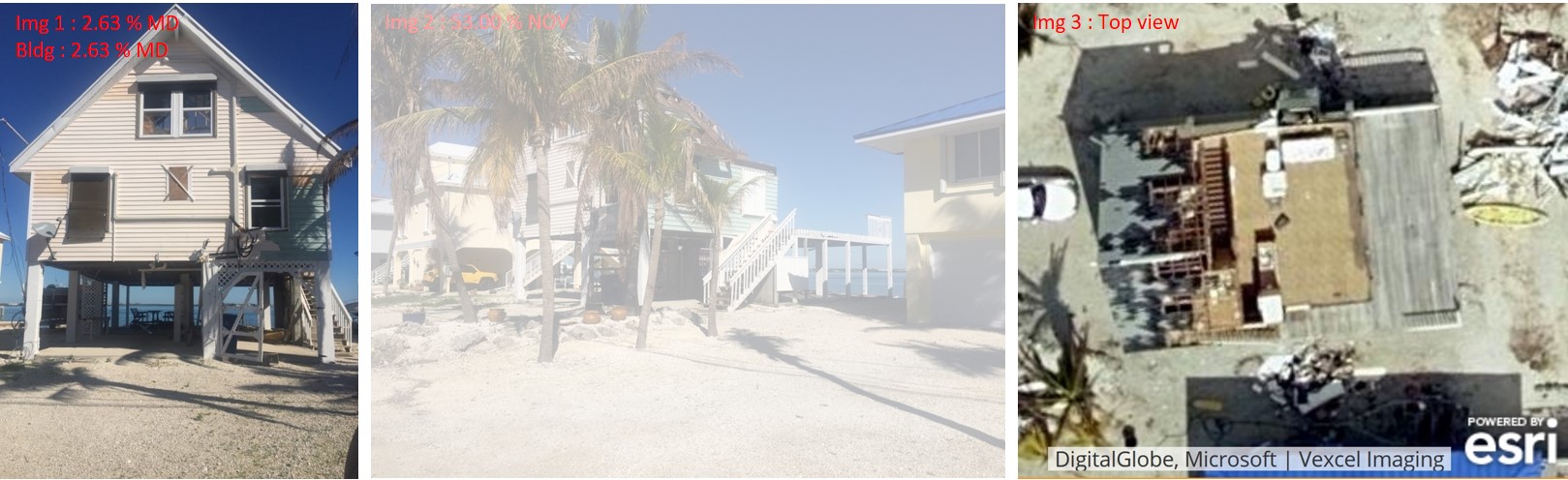}
    \caption{Sample of an incorrect building categorization.}
	\label{fig:sample_incorrect_MD}
\end{figure*}
To understand the limitations of the approach, it is informative to examine some specific building examples of correct (incorrect) decisions as well as ND.
Figure~\ref{fig:sample_res_MD} shows four images of a representative case in which a building is correctly categorized as MD. 
In this case, the first three raw images, numbered as 1, 2, and 3, do not show any evidence of damage. 
However, image number 4 does show the damage clearly, and the CNN classifies it as MD with a high probability. The fusion formula, Eq.~(\ref{eq:7}), categorizes the building as MD with high probability.

Figure~\ref{fig:sample_no_dec} includes six images corresponding to an ND case.
The true label of the building is MD. 
The three images in top row, numbered as 1, 2, and 3 are each individually classified as NOV with a high probability.
However, image number 4 does show signs of damage on the roof, albeit with a $51.79 \%$ probability.
Images 5 and 6 do not show any evidence of damage.
The fusion formula, also gives an almost fifty-fifty chance of MD.

Figure~\ref{fig:sample_incorrect_MD} corresponds to a case that is incorrectly categorized as NMD due to a shortage of informative images. 
In particular, there is not an adequate number of images to cover the building (remember that in this case study we have set $q_n = 1$, i.e., our framework mistakenly ``thinks'' that the building is covered).
Only one image (front view of the building facade, numbered 1) is classified as OV. 
Image number 2 shows canonical view of the building and potentially could capture the damage, but is highly obstructed by trees. 
Therefore, image 2 is classified as NOV and is not used for building categorization.
Thus, image number 1 is the only image available for detecting the overall damage condition which does not have any evidence that the building should be categorized as having major damage, and is not classified as damaged.
However, image number 3, which is the top view of the building capture through aerial imagery, which is not part of the data collected in preliminary survey, does show the damage on the back side of the building clearly. 
Note that this image would have been filtered out automatically by the overview classifier. 
It is included manually here for demonstrating the true building label. 
Investigating the case shown in \ref{fig:sample_incorrect_MD} reveals that the need for capturing multiple post-event images that cover all around the building is critical for correct building categorization, see Sec. \ref{tab:conf_setup2}.

\begin{table}
\caption{Confusion matrix using a loss function with parameters ($\alpha_1 = \alpha_2 = 0.3$) ; considering a sample coverage probability, $ q_1 =0.2, q_2 = 0.5, q_3 = 0.9, q_n =1$ for $ n \geq 4 $.}
\label{tab:conf_setup2}
\centering
\begin{tabular}{c c c c c c c}
&       &\multicolumn{3}{c}{Decision}\\
&      &  \cellcolor{gray!10}No OV & \cellcolor{gray!10}ND& \cellcolor{gray!10}MD & \cellcolor{gray!10}NMD & \cellcolor{gray!50}All\\
&\cellcolor{gray!10}No label& 26 & 19 & 5 & 4 & 54\\
&\cellcolor{gray!10}MD & 44  & \cellcolor{yellow!20}45 & \cellcolor{green!20}151 & \cellcolor{red!20}10 & 250\\
&\cellcolor{gray!10}NMD & 109 & \cellcolor{yellow!20}355 & \cellcolor{red!20}71 & \cellcolor{green!20}282 & 817\\
&\cellcolor{gray!50}All & 179 & 419 & 227 & 296 & 1,121\\
\rowcolor{gray!10} \cellcolor{white}
\rot{\rlap{\hspace{10pt}True label}}
\end{tabular}
\end{table}

\subsubsection{Discussion on selecting the coverage probability}
\label{sec:tunq}

\begin{figure*}
	\centering
	\subfloat[Fusion probability with coverage probability of $ q_1 =0.2, q_2 = 0.5, q_3 = 0.9, q_n =1$ for $ n \geq 4 $.
	\label{fig:fusprobq1}]{\includegraphics[scale=0.5]{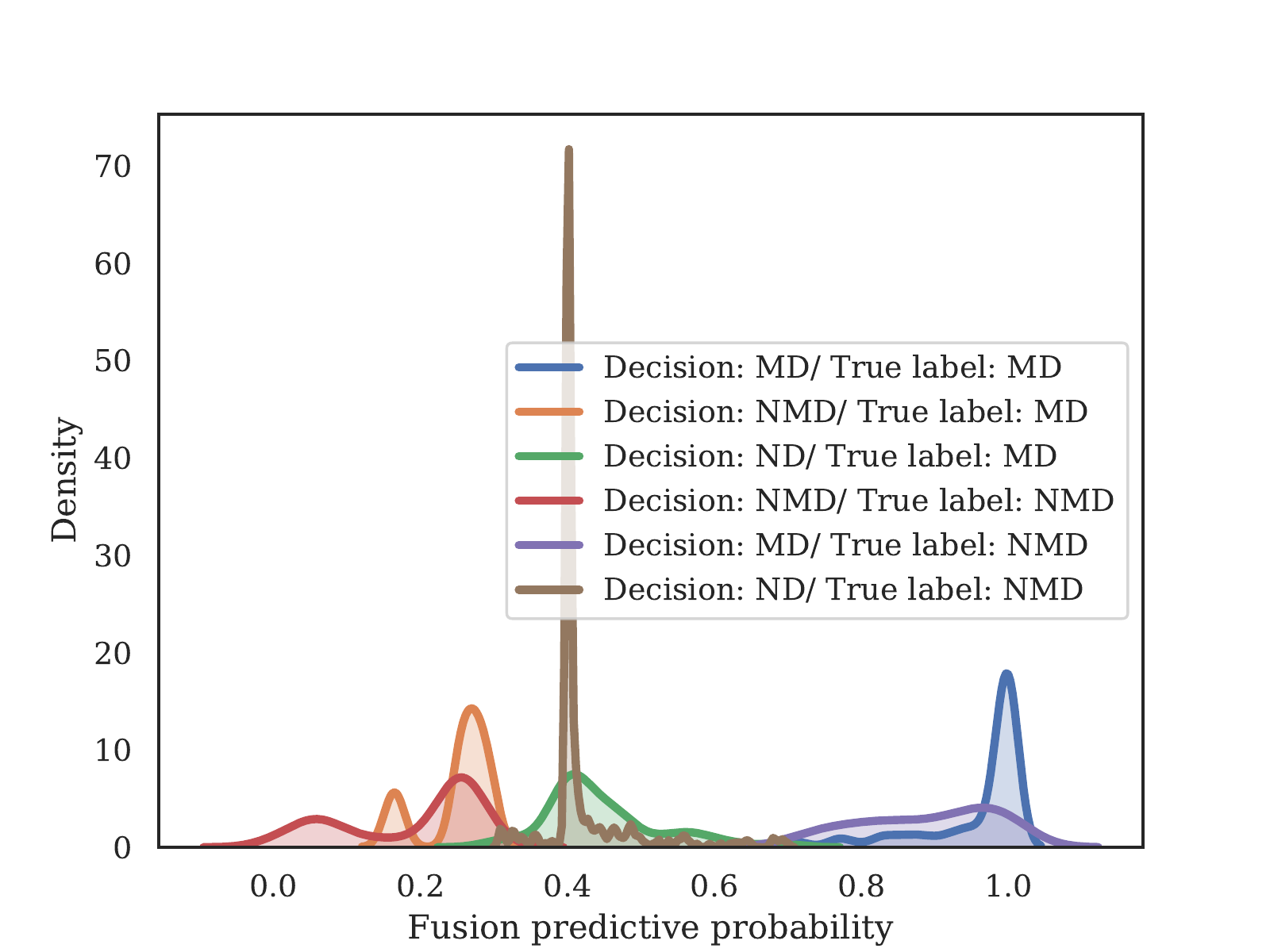}}
	\subfloat[Fusion probability with coverage probability of $ q_1 = q_2 = q_3 = 0, q_n =1$ for $ n \geq 4 $. 
	\label{fig:fusprobq2}]{\includegraphics[scale=0.5]{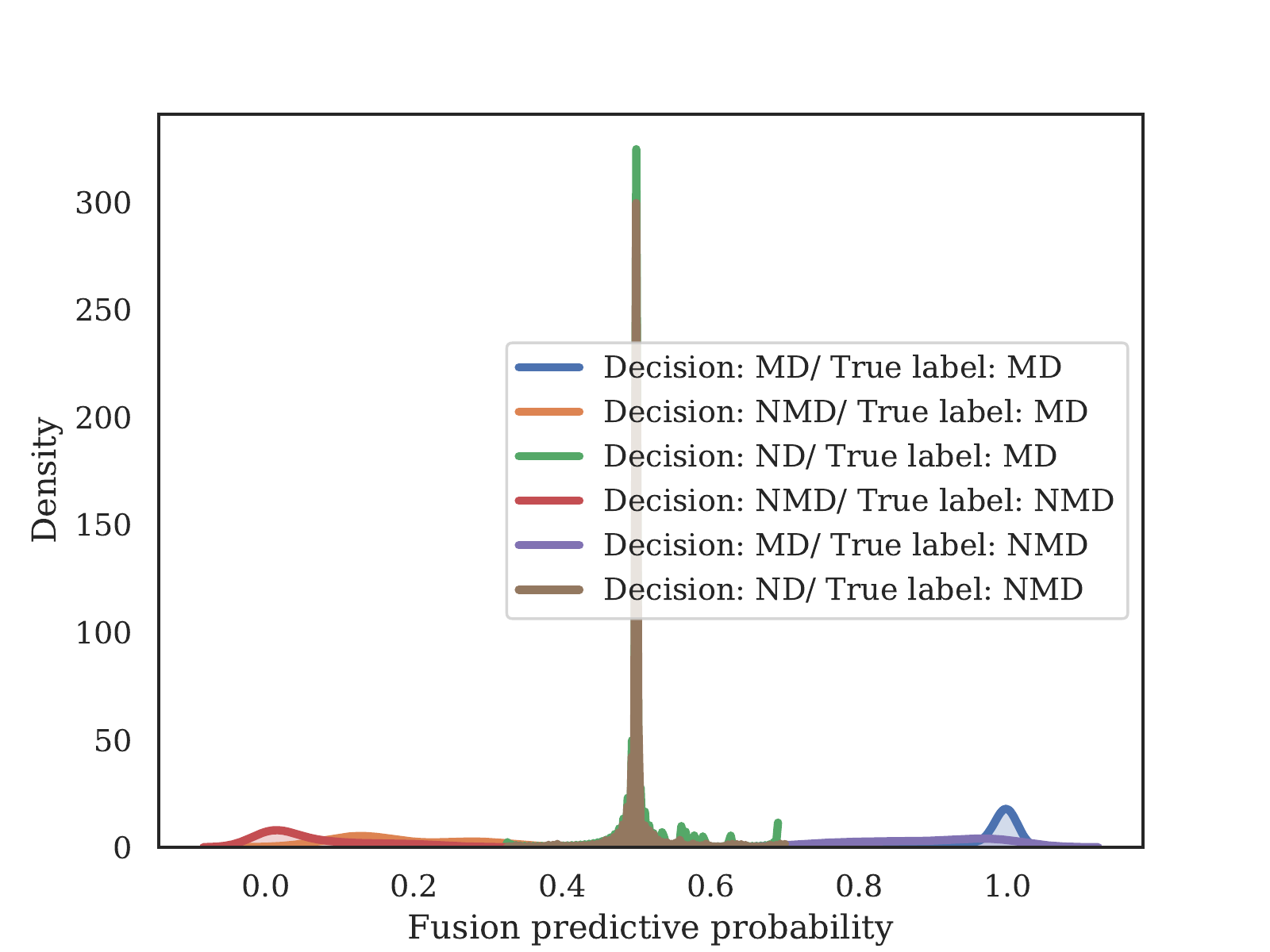}}
	\caption{Fusion probabilities.}
	\label{fig:fusprob}
\end{figure*}
The results presented in Table \ref{tab:conf_setup1} are based on the assumption that each given building is sufficiently covered, and human data collectors may have taken only 1 or 2 images of the NMD buildings.
However, our method is capable of dealing with unbiased data collected automatically.
This is possible through proper setup of $p(C=c|C_1=c_1,\dots,C_n=c_n)$, introduced in Sec.~ \ref{sec:post_info_fuse}. 
In Table~\ref{tab:conf_setup2} we illustrate the results of considering a sample coverage probability, $ q_1 =0.2, q_2 = 0.5, q_3 = 0.9, q_n =1$ for $ n \geq 4 $. 
The results in Table \ref{tab:conf_setup2} show that the number of MD buildings which are incorrectly characterized as NMD is reduced by almost 75\%, compared with Table \ref{tab:conf_setup1}.
These building are moved to the ND class.
For example, the case discussed in Fig. \ref{fig:ovnov} is characterized as ND after modifying the coverage probability.
Figure \ref{fig:fusprobq1} shows the density of the fusion predictive probabilities corresponding to different decision.
However, since one or two images are deemed insufficient to consider the building covered, the number of correctly detected NMD buildings also decreases by about 50\%, and again these are moved to the ND class. 
These consequences of incorporating coverage information can be interpreted as an indication that human data collectors typically have an inherent bias to take fewer images of buildings with no or minor damages, or NMD buildings.
The human collectors see things that are not depicted in the images they take.
For future utilization of this method, assuming the collected dataset contains more images of the target buildings, it is recommended to use realistic choice of coverage probability, e.g., $ q_1 = q_2 = q_3 = 0, q_n =1$ for $ n \geq 4 $. 
Density of the fusion predictive probabilities corresponding to different decisions are depicted in Fig. \ref{fig:fusprobq2} 

\subsubsection{Discussion on tuning the loss function}
\label{sec:tunlos}

\begin{figure*}
	\centering
	\subfloat[Loss function parameters on the accuracy.
	\label{fig:accall}]{\includegraphics[scale=0.5]{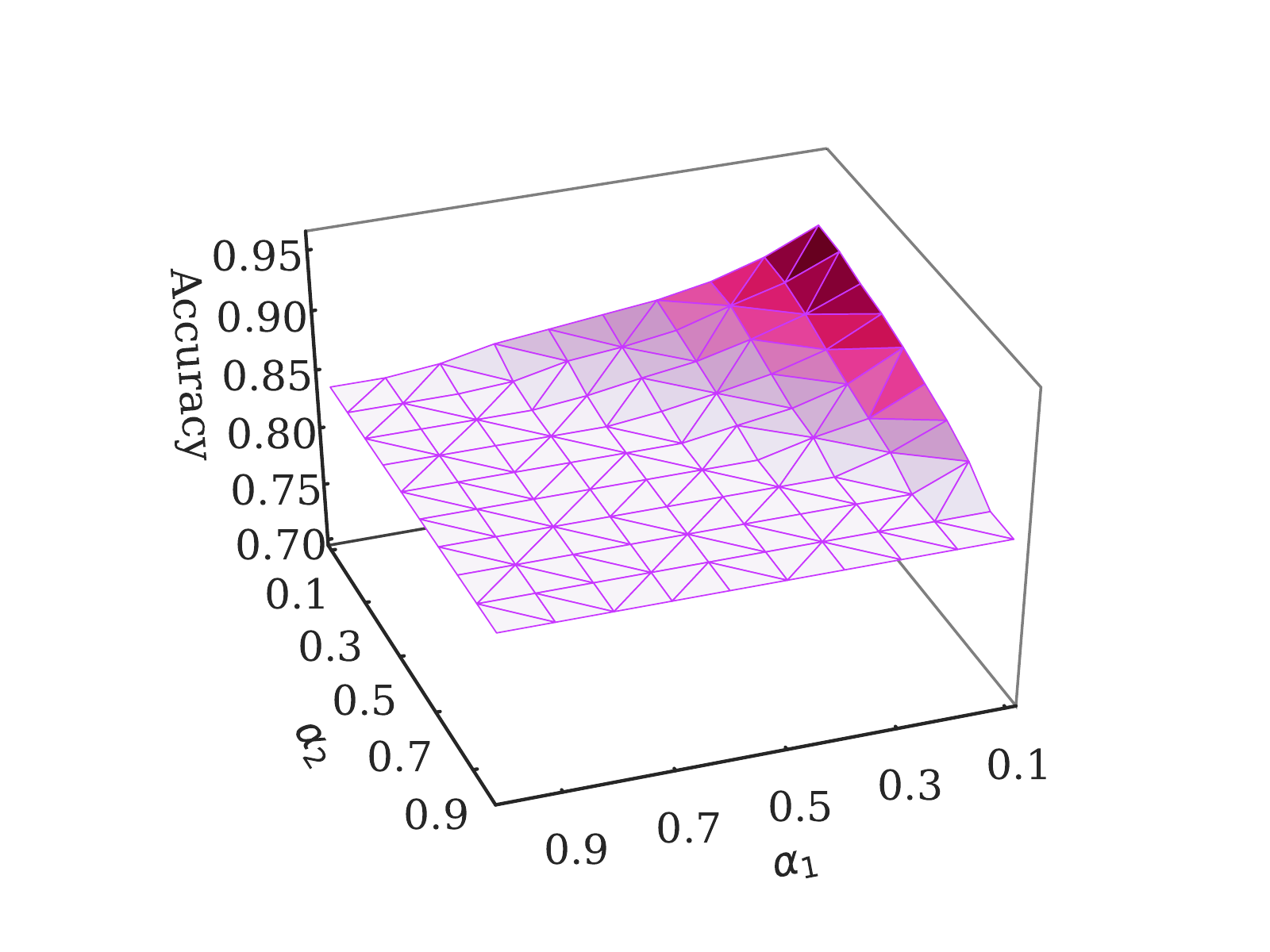}}
	\subfloat[Loss function parameters on the ND rate. \label{fig:ndrate}]{\includegraphics[scale=0.5]{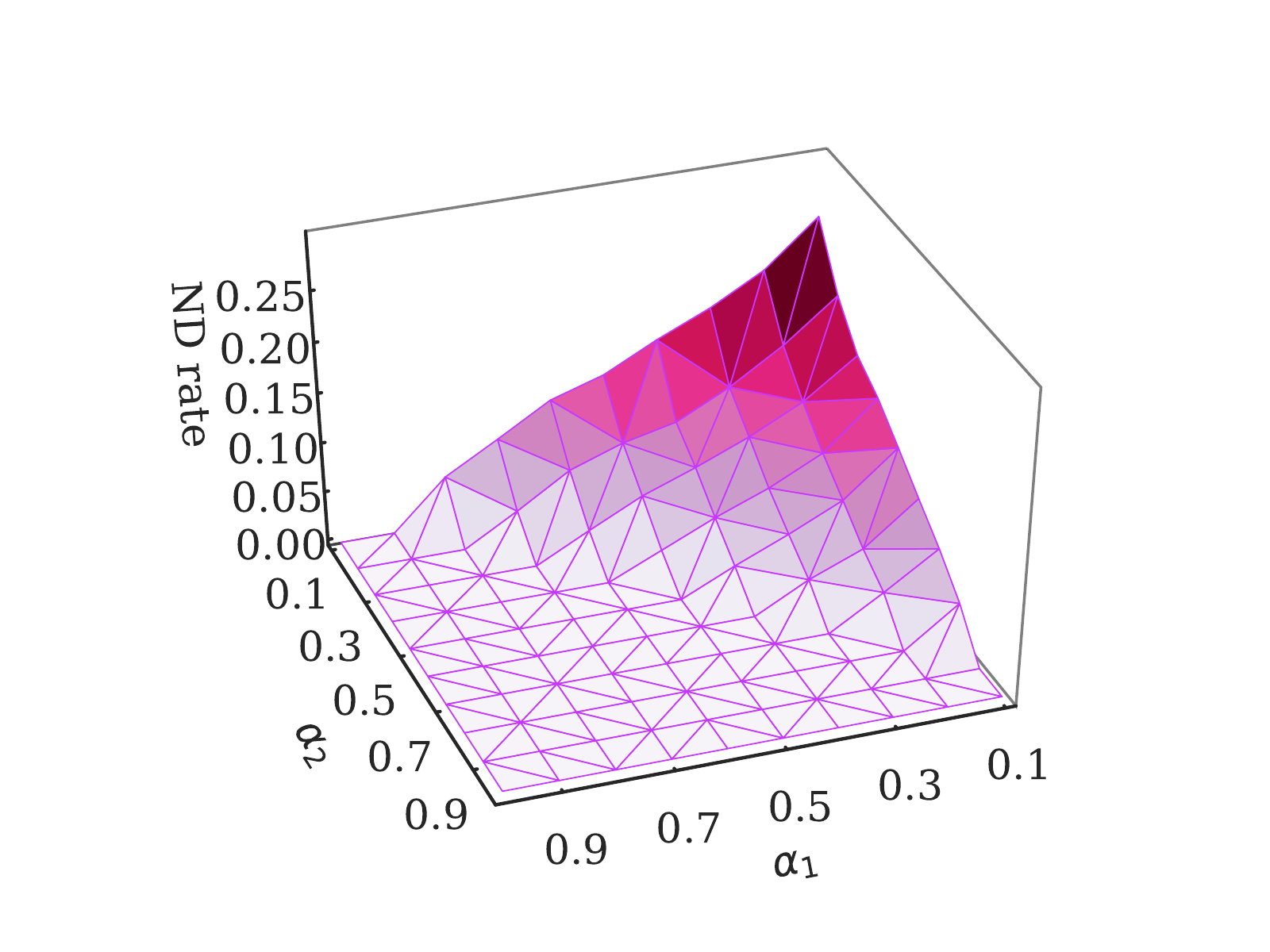}}
	\caption{Accuracy vs ND rate.}
	\label{fig:accvsnd}
\end{figure*}

In Tables \ref{tab:conf_setup1} and \ref{tab:conf_setup2}, the ratio of the correct, incorrect and ND prediction is highly dependent on the loss function parameters.
The choice of these parameters should reflect the objectives of the reconnaissance team.
To develop some intuition about these parameters, we investigate their effect on the results, we change $\alpha_1$ and $\alpha_2$ from 0.1 to 1 and calculate the results for all combination sets of the parameters. 
Figure \ref{fig:accall} demonstrates the effect of loss function parameters on the \emph{accuracy} of the post-event buildings overall damage categorization. 

According to, Fig. \ref{fig:accall}, decreasing both the parameters $\alpha_1$ and $\alpha_2$, results in higher accuracy.
However according to, Fig. \ref{fig:ndrate}, decreasing $\alpha_1$ and $\alpha_2$, results in a high \emph{ND rate}, rate of ND predictions over all permissible predictions.
To explain it more clearly, we describe two scenarios corresponding to two teams with different goals.
The first scenario refers to a team that has limited but sufficient resources to visit all potential MD buildings, and prefers to not miss any of the MD buildings.
In this scenario, high accuracy is not critical, albeit they want avoid a high ND rate which may lead to missing some MD cases.
They can encode this objective in the loss function by picking the $\alpha_1$ and $\alpha_2$ very high, e.g., 0.9.
The second scenario refers to a team that has a limited resources and prefers to spend it more conservatively and only visit the buildings that have high probability of falling into MD category.
In this scenario, the goal is to increase the accuracy, however, having high ND rate is not a big concern.
They can encode this objective by picking the $\alpha_1$ and $\alpha_2$ very small, e.g., 0.1.

\subsection{Pre-event stream validation}
\label{sec:mehtod_val_pre}

In the pre-event stream, images of 807 of the 1,121 buildings visited after hurricane Irma are successfully extracted from street view panoramas.
The 314 buildings excluded from the pre-event images extraction are not available because (1)  the building's address is not available, (2) the street view panoramas are not available, (3) the building facade maybe occluded by other objects, e.g., trees, cars or other buildings, (4) in some geographical regions street view images are not up to date and have a very low resolution. 
So we set our pre-event image extraction tool to filter out those images. 
Here, all of these 807 buildings are assumed to be captured adequately with the images available.

\begin{table}
\caption{Loss function.}
\label{tab:pre-loss_func}
\centering
\begin{tabular}{c c c c c c}
&       &\multicolumn{3}{c}{Decision}\\
&      &  \cellcolor{gray!10}ND& \cellcolor{gray!10}Attribute 1 & \cellcolor{gray!10}Attribute 2  \\
&\cellcolor{gray!10}Attribute 1  & $\alpha_1$  & 0 & 1 \\
&\cellcolor{gray!10}Attribute 2  & $\alpha_2$ & 1 & 0\\
\rowcolor{gray!10} \cellcolor{white}
\rot{\rlap{\hspace{10pt}True label}}
\end{tabular}
\end{table}
The general form of the loss function for determining pre-event attributes is shown in, Fig. \ref{tab:pre-loss_func}. Similar to the post-event loss function, the loss of a correct prediction is set to zero, but the loss of making mistakes, 1, or labeling as ND, $\alpha_1$ and $\alpha_2$, represents the relative penalties.

\begin{table}
\caption{Confusion matrix of first floor elevation using a loss function with parameters ($\alpha_1 = \alpha_2 = 0.3$).}
\label{tab:pre_conf_setup_fe}
\centering
\begin{tabular}{c c c c c c}
&       &\multicolumn{3}{c}{Decision}\\
&      &\cellcolor{gray!10}ND& \cellcolor{gray!10}Elevated & \cellcolor{gray!10}Not Elevated & \cellcolor{gray!50}All\\
&\cellcolor{gray!10}Elevated &\cellcolor{yellow!20}111 & \cellcolor{green!20}136 & \cellcolor{red!20}35 & 282\\
&\cellcolor{gray!10}Not Elevated &\cellcolor{yellow!20}143 & \cellcolor{red!20}20 & \cellcolor{green!20}362 & 525\\
&\cellcolor{gray!50}All &  254 & 156 & 397 & 807\\
\rowcolor{gray!10} \cellcolor{white}
\rot{\rlap{\hspace{10pt}True label}}
\end{tabular}
\end{table}

\begin{table}
\caption{Confusion matrix of number of stories using a loss function with parameters ($\alpha_1 = \alpha_2 = 0.3$).}
\label{tab:pre_conf_setup_ns}
\centering
\begin{tabular}{c c c c c c}
&       &\multicolumn{3}{c}{Decision}\\
&      &\cellcolor{gray!10}ND& \cellcolor{gray!10}One & \cellcolor{gray!10}Two & \cellcolor{gray!50}All\\
&\cellcolor{gray!10}One &\cellcolor{yellow!20}137 & \cellcolor{green!20}226 & \cellcolor{red!20}34 & 397\\
&\cellcolor{gray!10}Two &\cellcolor{yellow!20}67 & \cellcolor{red!20}19 & \cellcolor{green!20}209 & 295\\
&\cellcolor{gray!10}Unknown or more than Two &  16 & 12 & 87 & 115\\
&\cellcolor{gray!50}All &  220 & 257 & 330 & 807\\
\rowcolor{gray!10} \cellcolor{white}
\rot{\rlap{\hspace{10pt}True label}}
\end{tabular}
\end{table}

\begin{table}
\caption{Confusion matrix of construction material using a loss function with parameters ($\alpha_1 = \alpha_2 = 0.3$).}
\label{tab:pre_conf_setup_cm}
\centering
\begin{tabular}{c c c c c c}
&       &\multicolumn{3}{c}{Decision}\\
&      &\cellcolor{gray!10}ND& \cellcolor{gray!10} Masonry& \cellcolor{gray!10}Wood & \cellcolor{gray!50}All\\
&\cellcolor{gray!10}Masonry &\cellcolor{yellow!20}27 & \cellcolor{green!20}102 & \cellcolor{red!20}10 & 139\\
&\cellcolor{gray!10}Wood &\cellcolor{yellow!20}119 & \cellcolor{red!20}28 & \cellcolor{green!20}116 & 263\\
&\cellcolor{gray!10}Unknown or Others &  164 & 136 & 105 & 405\\
&\cellcolor{gray!50}All &  310 & 266 & 231 & 807\\
\rowcolor{gray!10} \cellcolor{white}
\rot{\rlap{\hspace{10pt}True label}}
\end{tabular}
\end{table}

Tables \ref{tab:pre_conf_setup_fe}, \ref{tab:pre_conf_setup_ns}, and \ref{tab:pre_conf_setup_cm} provide the confusion matrix for the results of our demonstration of the end-to-end, pre-event stream data analysis. 
These results are obtained with a loss function with $\alpha_1 = \alpha_2 = 0.3$.
Table \ref{tab:pre_conf_setup_fe} provides the confusion matrix for the results of our demonstration of the end-to-end, pre-event stream data analysis for first floor elevation attribute. 
Out of a total of 807 buildings, 498 buildings in the dataset posted are correctly categorized,  55 buildings are classified incorrectly, and 253 buildings labeled ND.
Table \ref{tab:pre_conf_setup_ns} provides the confusion matrix for the results for number of stories attribute.
Out of a total of 807 buildings, 115 buildings in the posted dataset have an unknown or more than two stories true label. 
Therefore data from the 692 one and two story labeled buildings are used here.
The results show that 435 buildings are correctly categorized, 53 buildings are classified incorrectly, and 204 buildings labeled ND.
Table \ref{tab:pre_conf_setup_cm} shows the confusion matrix for the results for construction material attribute. 
Out of a total of 807 buildings, 405 buildings have unknown or other types of material, and 402 buildings are labeled as either \emph{wood} or \emph{masonry} buildings.
Out of these 402 buildings, the automated data analysis procedure results show 218 buildings are correctly categorized, 38 buildings are classified incorrectly, and 146 buildings are labeled ND.

\section{Conclusion}
\label{sec:con}
After a natural disaster such as a hurricane, information about the performance of the built environment is gathered to learn lessons and to inform codes and guidelines.
A preliminary survey is conducted immediately after the event to identify the most valuable sites and buildings to visit during a more detailed survey that follows. 
That manual process is tedious and time consuming, but the strategic use of automation and computer vision can accelerate and even automate the process. 

In this paper, a technique is developed to directly support the needs of the human engineers conducting a preliminary survey. 
The technique is focused on automating the data analysis steps involved in this process, achieving this goal by leveraging and adapting recent advances in deep learning research to this important problem. 
The input to the technique is a collection of post-event images collected from residential buildings in the affected region. 
The output of the technique is the building attributes, and the damage classification for the buildings in that region. 
By formulating this data analysis problem in terms of a pre-event stream and a post-event stream, the critical information is automatically extracted from the images collected, for ready use by the human engineer. A classification schema is designed to organize the data. 
Robust scene classifiers are designed for specific scene classification tasks. 
Information fusion methods are developed to combine the results from multiple images, yielding a result that collectively considers the individual results of multiple images. 
Valuable lessons on how to achieve robust classification for such complex and unstructured datasets are also discussed. 
The technique is demonstrated using a publicly-available, real-world dataset collected by the NSF-funded StEER teams during the 2017 and 2018 hurricanes. 
The technique provides the engineer in the field with automated capabilities, reducing effort, improving consistency, and accelerating decisions after a major event. 
Because automation has enormous potential in the analysis of these images, the collection of more data, with less subjectivity, will make this process more robust and will also reduce bias in the results. 
Thus, collecting more data to learn from such events is strongly encouraged.
Future research that builds on this technique can be categorized into two major directions. 
The primary is direction is facilitating the collection and process of multiple sources of data, e.g., all type of images (street-level, aerial, and satellite), engineers' recorded and written observations, social media reports. 
Another direction is in generalizing the techniques to fuse the available types of information properly.

\section*{Acknowledgement}
The authors wish to acknowledge support from sources including: the  Center for Resilient Infrastructures, Systems, and Processes (CRISP) at Purdue, and the National Science Foundation under Grants No. NSF 1608762 and 1835473.

\bibliographystyle{cas-model2-names}

\bibliography{references}


\end{document}